\documentclass[lettersize,journal]{IEEEtran}
\usepackage{amsmath,amsfonts}
\usepackage{algorithmic}
\usepackage{algorithm}
\usepackage{array}
\usepackage[caption=false,font=normalsize,labelfont=sf,textfont=sf]{subfig}
\usepackage{textcomp}
\usepackage{stfloats}
\usepackage{url}
\usepackage{verbatim}
\usepackage{graphicx}
\usepackage{cite}
\usepackage{booktabs}
\usepackage{bbding}
\usepackage{threeparttable}
\usepackage{multirow}
\usepackage{orcidlink}

\usepackage{xurl}

\begin{document}

\title{More Clear, More Flexible, More Precise:\\ A Comprehensive  Oriented Object Detection benchmark for UAV}

\author{Kai Ye \orcidlink{0000-0003-0914-0213}, Haidi Tang \orcidlink{0009-0002-7771-2914}, Bowen Liu \orcidlink{0009-0000-5006-8925}, Pingyang Dai \orcidlink{0000-0001-9780-271X}, Liujuan Cao \orcidlink{0000-0002-7645-9606}, \textit{Member, IEEE}, Rongrong Ji \orcidlink{0000-0001-9163-2932}, \textit{Senior Member, IEEE} 
\thanks{ Kai Ye, Haidi Tang, Bowen Liu are with the Key Laboratory of Multimedia Trusted Perception
 and Efficient Computing, Ministry of Education of China, and the Institute
 of Artificial Intelligence, Xiamen University, Xiamen 361005, China (e-mail:
 yekai@stu.xmu.edu.cn; 23020231154225@stu.xmu.edu.cn; bowenliu@stu.xmu.edu.cn).

  Pingyang Dai is with the Key Laboratory of Multi
media Trusted Perception and Efficient Computing, Ministry of Education
 of China, Xiamen University, Xiamen 361005, China, and also with the
 School of Informatics, Xiamen University, Xiamen 361005, China (e-mail:
 pydai@xmu.edu.cn).

 Liujuan Cao is with the Key Laboratory of Multi
media Trusted Perception and Efficient Computing, Ministry of Education
 of China, Xiamen University, Xiamen 361005, China, and also with the
 School of Informatics, Xiamen University, Xiamen 361005, China (e-mail: caoliujuan@xmu.edu.cn).

Rongrong Ji is with the Key Laboratory of Multimedia Trusted Perception
 and Efficient Computing, Ministry of Education of China, and the Fujian
 Engineering Research Center of Trusted Artificial Intelligence Analysis and
 Application, Institute of Artificial Intelligence, Xiamen University, Xiamen
 361005, China, and also with the Peng Cheng Laboratory, Shenzhen 518066,
 China (e-mail: rrji@xmu.edu.cn).
 }}



\maketitle

\begin{abstract}
Applications of unmanned aerial vehicle (UAV) in logistics, agricultural automation, urban management, and emergency response are highly dependent on oriented object detection (OOD) to enhance visual perception. 
Although existing datasets for OOD in UAV provide valuable resources, they are often designed for specific downstream tasks.
Consequently, they exhibit limited generalization performance in real flight scenarios and fail to thoroughly demonstrate algorithm effectiveness in practical environments.
To bridge this critical gap, we introduce CODrone, a comprehensive oriented object detection dataset for UAVs that accurately reflects real-world conditions. 
It also serves as a new benchmark designed to align with downstream task requirements, ensuring greater applicability and robustness in UAV-based OOD.
Based on application requirements, we identify four key limitations in current UAV OOD datasets-low image resolution, limited object categories, single-view imaging, and restricted flight altitudes-and propose corresponding improvements to enhance their applicability and robustness.
Furthermore, CODrone contains a broad spectrum of annotated images collected from multiple cities under various lighting conditions, enhancing the realism of the benchmark. 
To rigorously evaluate CODrone as a new benchmark and gain deeper insights into the novel challenges it presents, we conduct a series of experiments based on 22 classical or SOTA methods.
Our evaluation not only assesses the effectiveness of CODrone in real-world scenarios but also highlights key bottlenecks and opportunities to advance OOD in UAV applications.
Overall, CODrone fills the data gap in OOD from UAV perspective and provides a benchmark with enhanced generalization capability, better aligning with practical applications and future algorithm development.
The dataset has been publicly released and is available at: \url{https://github.com/AHideoKuzeA/CODrone-A-Comprehensive-Oriented-Object-Detection-benchmark-for-UAV}.
\end{abstract}


\section{Introduction}
\IEEEPARstart{C}{ompared} to high-altitude remote sensing platforms such as satellites, UAVs or drones operate at lower altitudes with more flexible viewing angles, making them indispensable in urban applications such as agriculture, traffic planning, and logistics\cite{DBLP:journals/tgrs/JiangZWWZYZCC22,DBLP:journals/tgrs/WangGTGWGZH00CY24,li2024investigation,xu2023survey,guerin2021certifying}.
Among the core perception tasks that support UAVs in executing downstream missions, oriented object detection (OOD) \cite{DBLP:conf/cvpr/Luo00LY024, DBLP:conf/cvpr/Yu0LDD0Y24, DBLP:conf/cvpr/0010D23, DBLP:conf/cvpr/HuaLLLZYB23} plays a vital role.
As an extension of horizontal bounding box detection, OOD not only localizes and classifies objects but also estimates their rotation angles, providing more precise spatial information.
Due to the variable flight altitudes and diverse perspectives of UAVs, the application scenarios they face are inherently dynamic and complex.
This variability poses significant challenges for OOD models, making robust and generalized detection more difficult.

\begin{figure}
    \centering
    \includegraphics[width=0.9\linewidth]{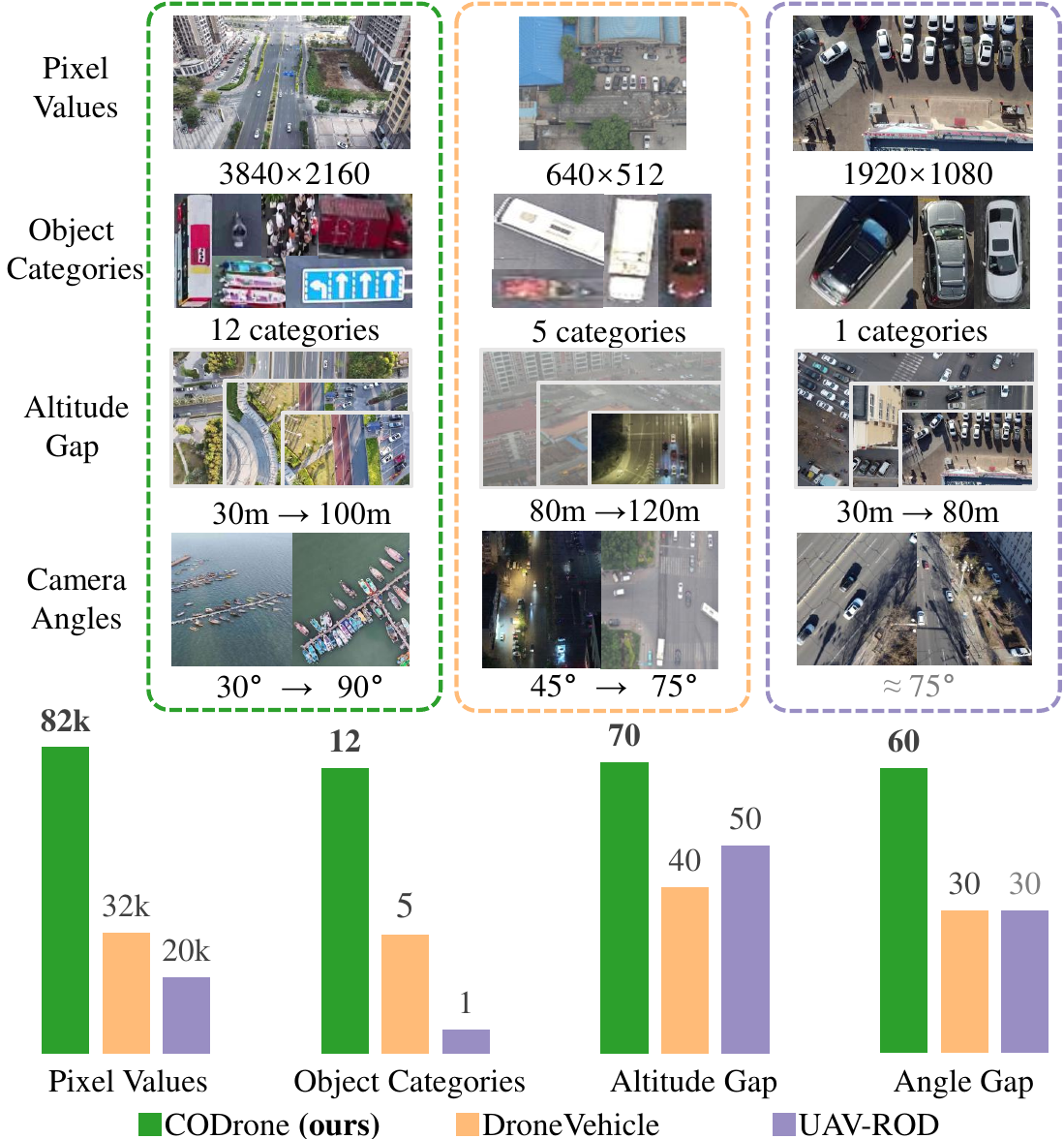}
    \caption{Comparison of the proposed CODrone dataset with existing UAV oriented object detection (OOD) datasets. 
    CODrone is designed as a UAV OOD benchmark for urban environments, constructed with high-resolution imagery, diverse object categories, and flexible variations in both altitude and camera angle. (The camera angle for UAV-ROD is not explicitly provided in the dataset and is therefore estimated based on available visual and metadata information.)}
    \label{fig:compare}
\end{figure}

Most existing remote sensing OOD datasets are primarily focused on high-altitude imaging platforms. 
For example, the SODA-A\cite{DBLP:journals/pami/ChengYYYZXH23} dataset emphasizes large-scale, high-resolution aerial imagery, driving advancements in algorithm performance under high-resolution airborne cameras. 
The DOTA\cite{DBLP:conf/cvpr/XiaBDZBLDPZ18} dataset is widely used in both industry and academia.
It provides a rich and challenging resource for training and evaluating aerial-oriented object detection models.
DIOR-R\cite{DBLP:journals/tgrs/ChengWLXLYH22} is a large-scale benchmark for oriented object detection in optical remote sensing imagery, covering 20 classes of landmark objects, such as stadiums and train stations. 
HRSC2016\cite{DBLP:conf/icpram/LiuYWY17}, a dataset focused on ship detection in remote sensing imagery, is particularly valuable due to the orientation-sensitive nature of ships, making it one of the standard benchmarks for evaluating OOD algorithms. 
Other datasets such as FGSD\cite{DBLP:journals/corr/abs-2003-06832}, VEDAI\cite{DBLP:journals/jvcir/RazakarivonyJ16}, and UCAS-AOD\cite{DBLP:conf/icip/ZhuCDFYJ15} have also contributed significantly by offering high-quality data for this domain.

\begin{figure}
    \centering
    \includegraphics[width=1\linewidth]{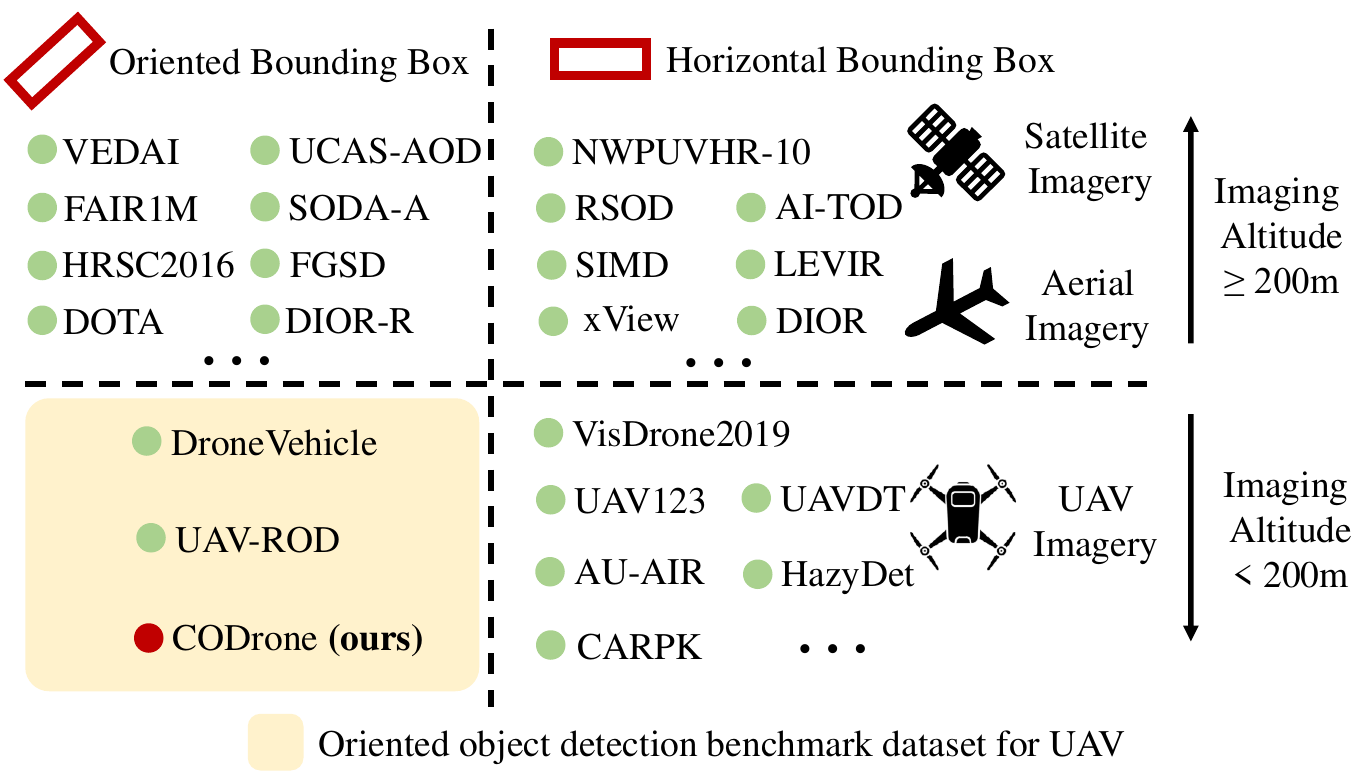}
    \caption{Classification of existing remote sensing object detection benchmark based on image source and annotation method. The proposed CODrone addresses the scarcity of oriented object detection benchmarks for UAVs, while also introducing new challenges in terms of object categories, viewing angles, and other key aspects.}
    \label{fig:differ}
\end{figure}

However, a key distinction between UAV imagery and the datasets collected by platforms like Google Earth and Gaofen lies in the much lower flight altitudes of UAVs. 
As a result, existing remote sensing datasets fail to fully represent the image characteristics encountered in real-world UAV scenarios.
Although UAV-based datasets such as VisDrone2019\cite{DBLP:conf/iccvw/DuZWWSLBSZKLZAS19}, AU-AIR\cite{DBLP:conf/icra/BozcanK20} , UAVDT\cite{DBLP:conf/eccv/DuQYYDLZHT18}, and CARPK\cite{DBLP:conf/iccv/HsiehLH17} provide extensive visual data from drone perspectives, they lack oriented annotations, making them unsuitable for addressing oriented object detectio tasks.
As illustrated in Fig.\ref{fig:differ}, there is a significant gap in both data availability and benchmark development for OOD in UAV-based scenarios.
Currently, the publicly available datasets in this domain, DroneVehicle\cite{DBLP:journals/tcsv/SunCZH22} and UAV-ROD\cite{DBLP:journals/ijon/ZhouFLHP22}, are both limited to vehicle categories, making them insufficient to meet the diverse requirements of real-world UAV applications.
Moreover, limitations in data collection and annotation quality further hinder progress in oriented object detection within the UAV field.

{\bf{1. Image resolution needs to be improved.}}
With the rapid advancement of UAV hardware, the resolution of onboard imaging systems has significantly increased. 
Image resolution directly determines the size of the captured images and the number of valid objects that can be detected. 
However, processing high-resolution images remains a major challenge in remote sensing OOD, requiring dedicated datasets to support algorithm development. 
At the same time, in low-resolution settings, small objects such as pedestrians and bicycles often appear too blurred to be reliably detected, resulting in poor performance of existing methods. 
Therefore, enhancing the resolution of dataset images serves two purposes: it facilitates the development of OOD algorithms tailored for high-resolution data, and it provides richer, clearer object features that improve the comprehensiveness of algorithm evaluation.

{\bf{2. Object categories need to be enriched.}}
The demand for UAV oriented object detection (OOD) in real-world applications is rapidly increasing. 
However, existing UAV OOD datasets are predominantly focused on vehicle categories, failing to cover the diverse range of objects encountered in practical scenarios. 
While horizontal bounding box (HBB) datasets such as VisDrone\cite{DBLP:conf/iccvw/DuZWWSLBSZKLZAS19} offer a broader object set, their lack of oriented bounding box (OBB) annotations makes them unsuitable for OOD tasks. 
On the other hand, datasets like DOTA\cite{DBLP:conf/cvpr/XiaBDZBLDPZ18} do provide OBB annotations, but their high-altitude imaging perspective differs significantly from UAV viewpoints, introducing a substantial domain gap that leads to degraded model performance when applied to UAV imagery. 
Consequently, a UAV-based OOD dataset with diverse and realistic object category annotations is essential to meet the demands of applications.

{\bf{3. Multi-altitude imaging needs to be expanded.}}
One of the key advantages of UAVs over other remote sensing platforms is their flexible flight altitude, enabling data acquisition across a wide range of altitudes. 
However, this also introduces a significant challenge for oriented object detection: the variation in data distribution caused by altitude differences. 
For instance, at low altitudes, objects may appear medium-sized or even large, whereas the same objects captured from higher altitudes become much smaller. 
This demands strong spatial generalization capabilities from detection algorithms. 
Existing datasets, however, are typically collected at fixed or narrowly varying altitudes, limiting their ability to support the development and evaluation of models that generalize well across altitude domains. 
As a result, current algorithms tend to perform optimally only at specific flight heights.

{\bf{4. Diverse viewing angles are required.}}
In aerial imaging, the appearance and features of objects are highly dependent on the camera angle.
Modern UAVs are typically equipped with gimbal cameras that support flexible adjustment of viewing angles through varying flight postures.
In contrast to satellite and high-altitude imagery, which are predominantly captured from a nadir (top-down) perspective, vertical imaging is less operationally convenient for UAVs and does not fully leverage their flexible viewpoint.
Moreover, different viewing angles provide distinct visual cues: while top-down views offer only limited overhead information, oblique views (e.g., at 30°) can reveal side profiles of objects, enriching spatial context. 
This variability poses a key challenge for oriented object detection (OOD) in UAV scenarios, requiring models to maintain robust detection and alignment of object features across varying perspectives.

To address the aforementioned limitations in existing UAV-oriented object detection datasets and benchmarks, we present CODrone, a comprehensive OOD benchmark dataset in UAV scenarios.
A comparison between CODrone and existing UAV OOD datasets is illustrated in Fig.\ref{fig:compare}.
Pixel Values indicate the image resolution; Object Categories refer to the number of annotated classes; Altitude Gap represents the range between the lowest and highest flight altitudes; and Camera Angles denote the diversity of image acquisition perspectives.
CODrone consists of over 10,000 high-resolution images captured from real UAV flights across five cities, covering a variety of urban and industrial environments, including ports and docks.
To improve robustness and generalization, it includes images from the same scenes under normal lighting, low light, and night conditions.
We adopted three flight altitudes and two commonly used camera angles, resulting in six distinct viewpoint configurations.

All images are annotated with oriented bounding boxes across 12 common object categories, totaling more than 590,000 labeled instances.
Overall, this work constructs a comprehensive dataset and benchmark for oriented object detection in urban UAV scenarios, aiming to meet both research and practical application needs in the field.
The main contributions of this work are summarized as follows:

\begin{itemize}
\item{We proposed a large-scale, high-resolution UAV-oriented object detection dataset, CODrone, which consists of over ten thousand UAV-captured images with precise oriented bounding box annotations.
To the best of our knowledge,  CODrone is the first UAV object detection dataset featuring oriented annotations across a wide range of object categories.}
\item{The proposed CODrone dataset considers multiple influential factors, including image acquisition altitude, camera perspective, lighting conditions, and geographic location. These considerations enhance the dataset’s ability to support robust model training and realistic performance evaluation for UAV-based oriented object detection.}

\item{Based on the proposed dataset, we establish a UAV-oriented object detection benchmark and conduct training and evaluation using 22 representative or state-of-the-art methods. 
We identify existing limitations and challenges in the field, which can support future advancements in both algorithm development and practical deployment.}

\end{itemize}

\section{RELATED WORKS}
In UAV-based oriented object detection, particularly for data-driven deep learning approaches, datasets are indispensable.
To gain a comprehensive understanding of existing remote sensing object detection datasets, we categorize representative datasets into four main types based on sensing platforms and annotation methods as illustrated in Fig.\ref{fig:differ}.
For clarity, we define high-altitude platforms as those with sensor heights exceeding 200 meters, which exceed the operational range of most small UAVs, while images acquired at altitudes below 200 meters using actual UAVs are designated as UAV-based datasets.
We review a broad spectrum of datasets ranging from high-altitude horizontal object detection (HOD), high-altitude oriented object detection (OOD), and low-altitude UAV-based HOD, to the UAV-based OOD which is the primary focus of this work.

\subsection{High-altitude HOD dataset}
{\bf SIMD}\cite{DBLP:journals/staeors/HaroonSF20} comprises 5,000 images sourced from Google Earth at a resolution of 1024 × 768 pixels. 
It includes 15 object categories, such as seven types of vehicles, six types of aircraft, ships, and other classes. 
{\bf COWC}\cite{DBLP:conf/eccv/MundhenkKSB16} focuses specifically on vehicle detection and counting using horizontal bounding boxes. 
Its images were collected from six geographically diverse locations, and it includes 58,000 hard negative samples to improve generalization.
Meanwhile, {\bf NWPU VHR-10}\cite{DBLP:journals/tip/ChengHZX19} contains 800 very high-resolution images covering 10 geo-spatial object categories.
Among them, 650 images include annotated objects and 150 are background images. 
{\bf RSOD}\cite{xiao2015elliptic} provides annotations for only four categories: airplanes, oil tanks, sports fields, and overpasses in PASCAL VOC format. 
{\bf DIOR}\cite{DBLP:journals/corr/abs-1909-00133} further expands the scale of object detection benchmarks by providing over 20,000 optical remote sensing images and 190,000 object instances across 20 categories.
{\bf xView}\cite{DBLP:journals/corr/abs-1802-07856}, another large-scale benchmark, comprises more than 1 million annotated objects from 60 categories.
These satellite images were captured by the WorldView-3 satellite at distance of 0.3-meter ground sampling and span an area of more than 1,400 square kilometers.

{\bf HRRSD}\cite{DBLP:journals/tgrs/ZhangYFL19} includes 21,761 high-resolution images ranging from 0.15 to 1.2 meters in spatial resolution, collected from both Google Earth and Baidu Map. It contains 55,740 object instances across 13 categories, offering a diverse set of scenes and scales for remote sensing object detection research.
{\bf LEVIR}\cite{DBLP:journals/tip/ZouS18}
covers a wide range of terrestrial features commonly found in human living environments, such as urban areas, rural regions, mountainous zones, and coastal areas. Extreme landforms like glaciers, deserts, and gobi deserts are not included. The dataset contains three object categories: airplanes, ships (including offshore and seagoing vessels), and oil tanks.
{\bf AI-TOD}\cite{DBLP:conf/icpr/WangYGZX20}
is a large-scale dataset focused on the detection of small objects in aerial images, designed to promote advancements in small object detection techniques. Compared to other aerial object detection datasets, the average object size in AI-TOD is significantly smaller, approximately 12.8 pixels.

\subsection{High-altitude OOD dataset}
Among existing oriented object detection datasets in high-altitude remote sensing, {\bf DOTA} \cite{DBLP:conf/cvpr/XiaBDZBLDPZ18} is one of the most widely used.
It currently includes three versions, with the latest version, DOTA-v2, expanding the number of object categories to 18 and containing 11,268 images and 1,793,658 object instances.
{\bf SODA-A}\cite{DBLP:journals/pami/ChengYYYZXH23} is a high-resolution aerial image subset of the small object detection dataset SODA\cite{DBLP:journals/pami/ChengYYYZXH23}, with images collected from hundreds of cities worldwide, providing rich data diversity.
{\bf DIOR-R}\cite{DBLP:journals/tgrs/ChengWLXLYH22}, an extended version of the DIOR\cite{DBLP:journals/corr/abs-1909-00133} dataset, includes oriented bounding box annotations for 20 object categories, making it the OOD dataset with the largest number of annotated classes (excluding fine-grained subclasses). 
{\bf FAIR1M} \cite{DBLP:journals/corr/abs-2103-05569} introduces fine-grained categorization by splitting broad categories into subcategories, such as dividing airplanes into 11 sub-classes and vehicles into 10 subclasses, thus presenting a greater challenge for fine-grained OOD tasks.

{\bf HRSC2016}\cite{DBLP:conf/icpram/LiuYWY17} is a domain-specific dataset focused on ship detection, offering multi-level class for different ship types.
Meanwhile, {\bf VEDAI}\cite{DBLP:journals/jvcir/RazakarivonyJ16} contains a variety of object types such as vehicles and ships, with diverse image resolutions ranging from low to high, covering a wide spectrum of scenarios.
{\bf FGSD}\cite{DBLP:journals/corr/abs-2003-06832} is also a fine-grained ship detection dataset in high-resolution satellite imagery. 
It contains over 15,000 images with sub-meter resolution and more than 1 million annotated ship instances collected from major global ports, covering over a hundred cities, towns, airports, and harbors.
{\bf EAGLE}\cite{DBLP:conf/icpr/AzimiBHK20} 
is a large-scale remote sensing dataset designed for multi-class vehicle detection, containing 215,986 vehicle instances with orientation annotations. The images are captured from diverse real-world scenarios, encompassing various sensors, resolutions, flight altitudes, weather conditions, and lighting environments.
Finally, {\bf UCAS-AOD} \cite{DBLP:conf/icip/ZhuCDFYJ15} is known for its high-resolution aerial images and precise object annotations. 
It offers rich visual diversity across various scenes and viewpoints and focuses specifically on the detection of vehicles and aircraft—two highly relevant object categories in aerial imagery.

\begin{table*}[htbp]
  \centering
  \caption{Comparison between CODrone and existing commonly used UAV-based object detection benchmark datasets.}
  \begin{threeparttable}
    \begin{tabular}{c|r@{\,}c@{\,}lcccr@{\,}l r@{\,}lc}
    \toprule
    Dataset & \multicolumn{3}{c}{Resolution} & Categories & Altitude Gap & Camera Angles & \multicolumn{2}{c}{Images} & \multicolumn{2}{c}{Objects} & OBB \\
    \midrule
    VisDrone2019\cite{DBLP:conf/iccvw/DuZWWSLBSZKLZAS19} & 2000  & ×     & 1500  & 10    & *     & *     & 10.2  & k     & 54.2  & k     & \XSolid \\
    UAVDT\cite{DBLP:conf/eccv/DuQYYDLZHT18} & 1080  & ×     & 540   & 3     & 60m   & *     & 80.0  & k     & 841.5 & k     & \XSolid \\
    AU-AIR\cite{DBLP:conf/icra/BozcanK20} & 1920  & ×     & 1080  & 8     & 25m   & 45°    & 3.2   & k     & 132.0 & k     & \XSolid \\
    CARPK\cite{DBLP:conf/iccv/HsiehLH17} & 1280  & ×     & 720   & 1     & *     & *     & 1.4   & k     & 89.7  & k     & \XSolid \\
    HazyDet\cite{DBLP:journals/corr/abs-2409-19833} & 1333  & ×     & 800   & 3     & *     & *     & 11.6  & k     & 383.0 & k     & \XSolid \\
    \midrule
    DroneVehicle\cite{DBLP:journals/tcsv/SunCZH22} & 840   & ×     & 712   & 5     & 40m   & 30°    & 56.8  & k     & 953.0 & k     & \Checkmark \\
    UAV-ROD\cite{DBLP:journals/ijon/ZhouFLHP22} & 1920  & ×     & 1080  & 1     & 50m   & *     & 1.5   & k     & 30.0  & k     & \Checkmark \\
    \bf{CODrone (ours)} & \bf{3840}  & \bf{×}     & \bf{2160}  & \bf{12}    & \bf{70m}   & \bf{60°}    & 10.0  & k     & 596.7 & k     & \Checkmark \\
    \bottomrule
    \end{tabular}%
  \label{tab:da}%
   \begin{tablenotes}
        \footnotesize
        \item * indicate that the corresponding information is either not explicitly provided or not annotated in datasets.
      \end{tablenotes}
  \end{threeparttable}
\end{table*}%

\subsection{UAV-based HOD dataset}
The scarcity of UAV datasets is primarily due to the high cost of image acquisition, which requires real UAV equipment and manual data collection in specific environments.
As HOD is one of the fundamental perception tasks for UAVs, in recent years researchers have been increasing efforts to alleviate the lack of UAV-specific data through dedicated data collection and annotation.
Among them, {\bf VisDrone} \cite{DBLP:conf/iccvw/DuZWWSLBSZKLZAS19} is a widely used benchmark and international challenge for UAV vision, offering large-scale drone imagery captured across diverse scenes, weather conditions, and time periods. 
{\bf UAVDT}\cite{DBLP:conf/eccv/DuQYYDLZHT18} focuses on dynamic UAV views in complex traffic scenes and provides detailed annotations for challenges such as occlusion and varying viewpoints.

In addition, {\bf AU-AIR}\cite{DBLP:conf/icra/BozcanK20} combines UAV video with multi-modal sensor data, supporting research on multi-modal fusion in UAV vision tasks.
{\bf CARPK}\cite{DBLP:conf/iccv/HsiehLH17} provides nearly 90,000 annotated cars collected from four different parking lots using UAVs flying at approximately 40 meters altitude. 
It also includes annotations from selected images in the PUCPR\cite{DBLP:conf/iccv/HsiehLH17} dataset, totaling around 17,000 cars.
{\bf HazyDet}\cite{DBLP:journals/corr/abs-2409-19833} is a specialized dataset for UAV-based object detection in hazy conditions.
It integrates depth estimation and atmospheric scattering models to ensure realism and diversity, aiming to improve UAV perception under challenging environmental conditions.

\subsection{UAV-based OOD dataset}
While numerous high-quality datasets exist for high-altitude OOD and UAV-based HOD, publicly available datasets for UAV-based OOD remain extremely limited. 
Moreover, the existing UAV OOD datasets mainly focus on vehicle categories, which restricts their practical utility and the diversity of challenges they pose.
{\bf DroneVehicle}\cite{DBLP:journals/tcsv/SunCZH22} is a subtask of the VisDrone challenge, built on a mixed dataset of RGB and infrared images captured by UAVs. 
It is annotated with oriented bounding boxes for vehicles categorized into five distinct classes.
{\bf UAV-ROD}\cite{DBLP:journals/ijon/ZhouFLHP22} is a large-scale, high-resolution dataset collected from UAV platforms across multiple scenarios, also focused on vehicle detection.
It is particularly suitable for studying vehicle behavior in complex traffic scenes.
To the best of our knowledge, these are the only two publicly available benchmark datasets specifically designed for UAV-based OOD.

\section{DATASET CONSTRUCTION}
To address the scarcity of benchmarks and datasets in the domain of UAV-based oriented object detection, we construct a comprehensive oriented object detection dataset for UAVs, referred to as CODrone.
In this section, we provide a detailed overview of this new benchmark dataset, including its construction process, core characteristics, and the unique challenges it introduces to the field.


\subsection{Overview and comparison of datasets}
In Table \ref{tab:da}, we present a comparison between CODrone and other commonly used UAV-based object detection datasets.
It is important to note that datasets designed for more specialized tasks, such as multiview object detection, video-based dynamic object tracking, and other fine-grained UAV applications, are not included in the table, as their objectives and data structures differ significantly from standard image-based object detection tasks.
Similarly, datasets based on high-altitude remote sensing platforms fall outside the scope of this study and are therefore included as supplementary material in the appendix.

CODrone significantly expands several key dimensions, including image resolution, object category diversity, and variation in flight altitude and camera angle.
For resolution, CODrone employs a 3840 × 2160 high-resolution onboard camera, aligning with the capabilities of modern UAV hardware.
In terms of object classes, unlike most existing UAV OOD datasets that focus primarily on vehicles, CODrone includes a more diverse range of categories, thereby increasing the difficulty and realism of the detection task.
Furthermore, we explicitly annotate both altitude and camera angle for each image, enabling research into UAV pose-aware perception and related tasks. 
Although we collected a large amount of video data, we manually selected and annotated only 10,004 high-quality images for the OOD task. 
The remaining video resources will be labeled and released in future work to support related tasks such as UAV-based video object tracking and multi-frame detection.
\begin{figure*}
    \centering
    \includegraphics[width=0.95\linewidth]{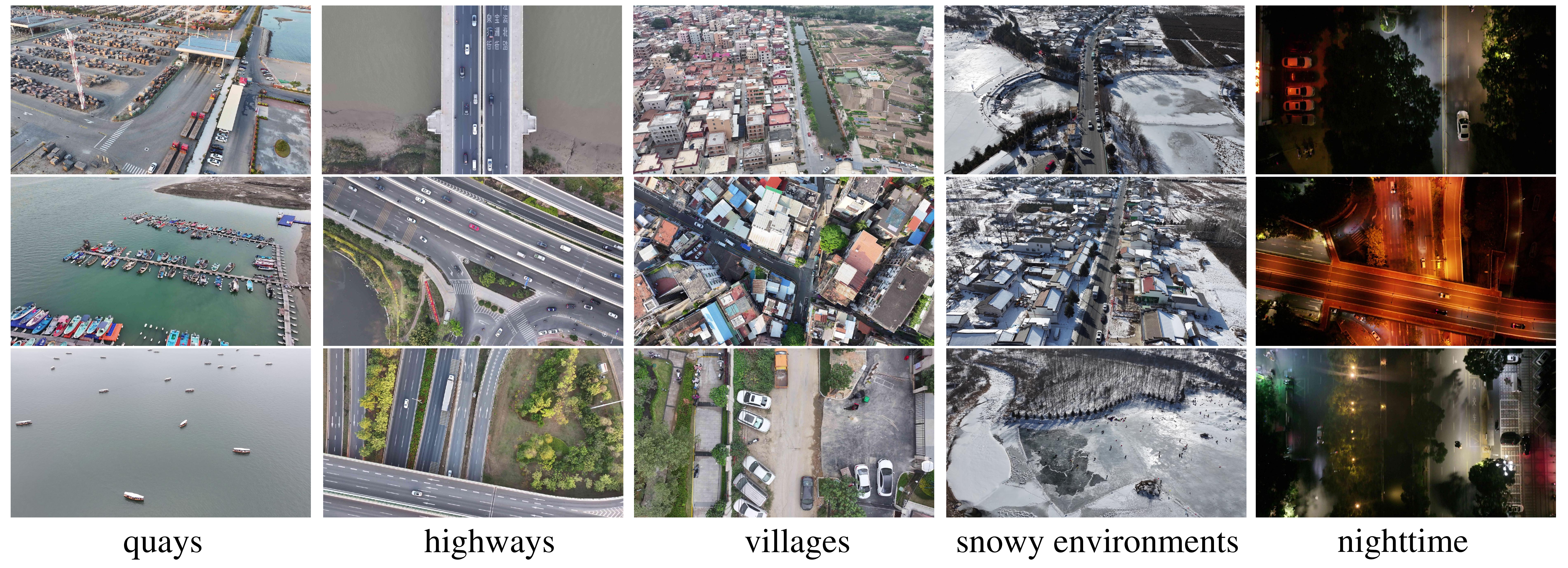}
    \caption{Visualization of representative image examples. The scenes shown above, such as quays, highways, villages, snowy environments, and nighttime conditions, are rarely observed in existing OOD UAV datasets.
    CODrone includes, but is not limited to, these diverse environments, thereby offering stronger general ization capabilities.}
    \label{fig:V1}
\end{figure*}

\subsection{Data capturing}


The CODrone was collected using a DJI Mavic 3 Pro UAV, flying at speeds ranging from approximately 10 to 15 meters per second.
The UAV was configured to capture imagery from two camera angles (30° and 90°) and at three flight altitudes (30 m, 60 m, and 100 m), resulting in a total of six unique viewpoint combinations.
This variation in acquisition perspectives makes the dataset more representative of the flexible and dynamic viewpoints encountered in real UAV operations.
As shown in Fig. \ref{fig:V1}, images were collected across a wide range of environments, including parks, urban streets, rural roads, bridges, parking lots, docks, and coastal areas, spanning over ten different regions.
To enhance the robustness and generalizability of the dataset, images were captured under various lighting conditions and weather scenarios, such as sunny, cloudy, daytime, and nighttime scenes.

We recorded hundreds of video clips at a frame rate of 60 frames per second (fps).
One frame was extracted every 200 frames, and we manually filtered out frames with noticeable motion blur to ensure that all retained images were both clear and contained visible objects.
In addition, redundant consecutive frames with nearly identical content were removed to preserve data diversity. 
As a result, we obtained a final set of 10,004 high-quality images.
Since the UAV operates at relatively high altitudes and the object sizes are small, no personally identifiable information is captured, thereby ensuring data privacy and security.

\begin{figure*}
    \centering
    \includegraphics[width=0.95\linewidth]{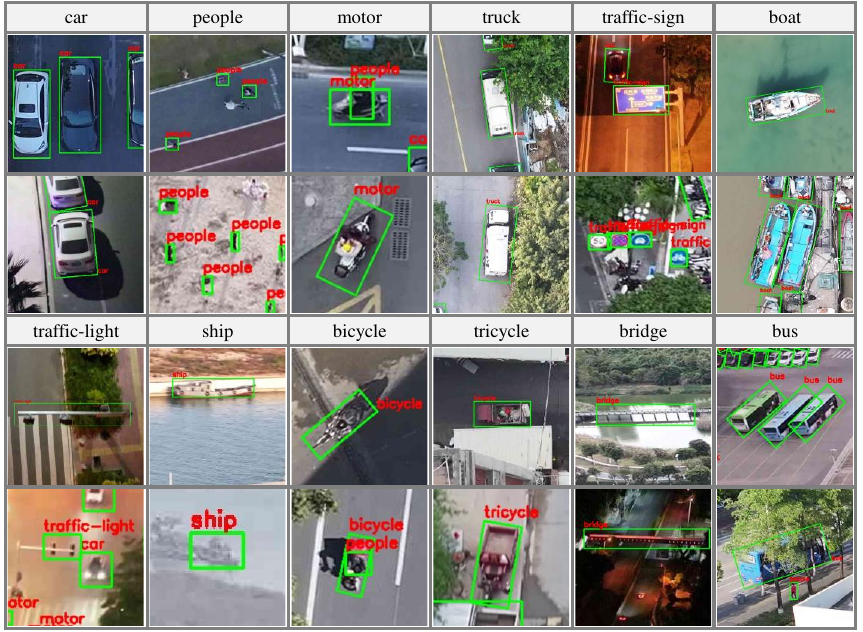}
    \caption{Visualization of annotated object categories in CODrone.
    We selected and labeled a wide range of object classes that are commonly encountered in urban environments and hold practical relevance for UAV applications.
    All instances are manually annotated with high-quality oriented bounding boxes.}
    \label{fig:class}
\end{figure*}

\subsection{Data labeling}



Given CODrone's intended application in urban UAV scenarios, we selected 12 object categories for annotation: car, truck, traffic-sign, people, motor, bicycle, traffic-light, tricycle, bridge, bus, boat, and ship. 
Notably, the bridge category refers specifically to pedestrian overpasses, not vehicular bridges.
These categories cover the most common objects observed in UAV-based urban perception tasks, ensuring high practical relevance.
A total of 443,592 object instances were annotated in the CODrone dataset. 
Fig. \ref{fig:class} illustrates representative samples for each category, while Table \ref{tab:class} provides a detailed breakdown of the number of annotated instances per class.

\begin{table}[htbp]
  \centering
  \caption{The distribution of instances for each category}
  \resizebox{0.45\textwidth}{!}{
    \begin{tabular}{c|cccc}
    \toprule
    Instances & Total  & Train  &  Val  &  Test \\
    \midrule
    car   & 227751 & 112588 & 46396 & 68767 \\
    truck  & 24431 & 12147 & 5058  & 7226 \\
    traffic-sign & 11797 & 5706  & 2425  & 3666 \\
    people & 79485 & 39343 & 15457 & 24685 \\
    motor & 73593 & 36662 & 14986 & 21945 \\
    bicycle & 3835  & 1892  & 813   & 1130 \\
    traffic-light & 6891  & 3336  & 1498  & 2057 \\
    tricycle & 2845  & 1398  & 599   & 848 \\
    bridge & 408   & 180   & 77    & 151 \\
    bus   & 4979  & 2444  & 1021  & 1514 \\
    boat  & 7143  & 3607  & 1080  & 2456 \\
    ship  & 434   & 224   & 58    & 152 \\
    ignored  & 16804 & 8169  & 3590  & 5045 \\
    \bottomrule
    \end{tabular}%
    }
  \label{tab:class}%
\end{table}%

Following the approach used in SODA\cite{DBLP:journals/pami/ChengYYYZXH23}, densely packed objects that could not be individually labeled were annotated as ignored. 
These ignored bounding boxes are excluded from training, helping to distinguish foreground objects from background noise and enhancing annotation precision.
In addition, we explicitly annotated partially occluded objects but discarded those with an occlusion ratio exceeding 80\% of the object area, in order to strengthen the robustness in handling partially visible objects.

Annotations in CODrone follow a triplet format ${L, C, D}$, where $L$ represents the location of object, $C$ is the category, and $D$ is a binary difficulty indicator. 
For oriented bounding box annotations, $L$ is represented by the coordinates of the four corner points: $L=\{(x_i, y_i)\}_{i=1}^4$.
This format is consistent with existing OOD datasets. 
The difficulty label $D$ is assigned as 1 for hard samples, including small objects (smaller than 32×32 pixels), heavily occluded objects (occlusion ratio $\geq$ 0.5), and blurred objects due to extreme perspective distortion.
All other objects are considered normal and assigned a $D$ value of 0.

In line with previous studies, CODrone is split into three subsets: training (50\%), validation (20\%), and testing (30\%), with all annotations for the test set publicly available to facilitate reproducible evaluation.

\subsection{Characteristics and challenges of the dataset}
\begin{figure*}
    \centering
    \includegraphics[width=0.9\linewidth]{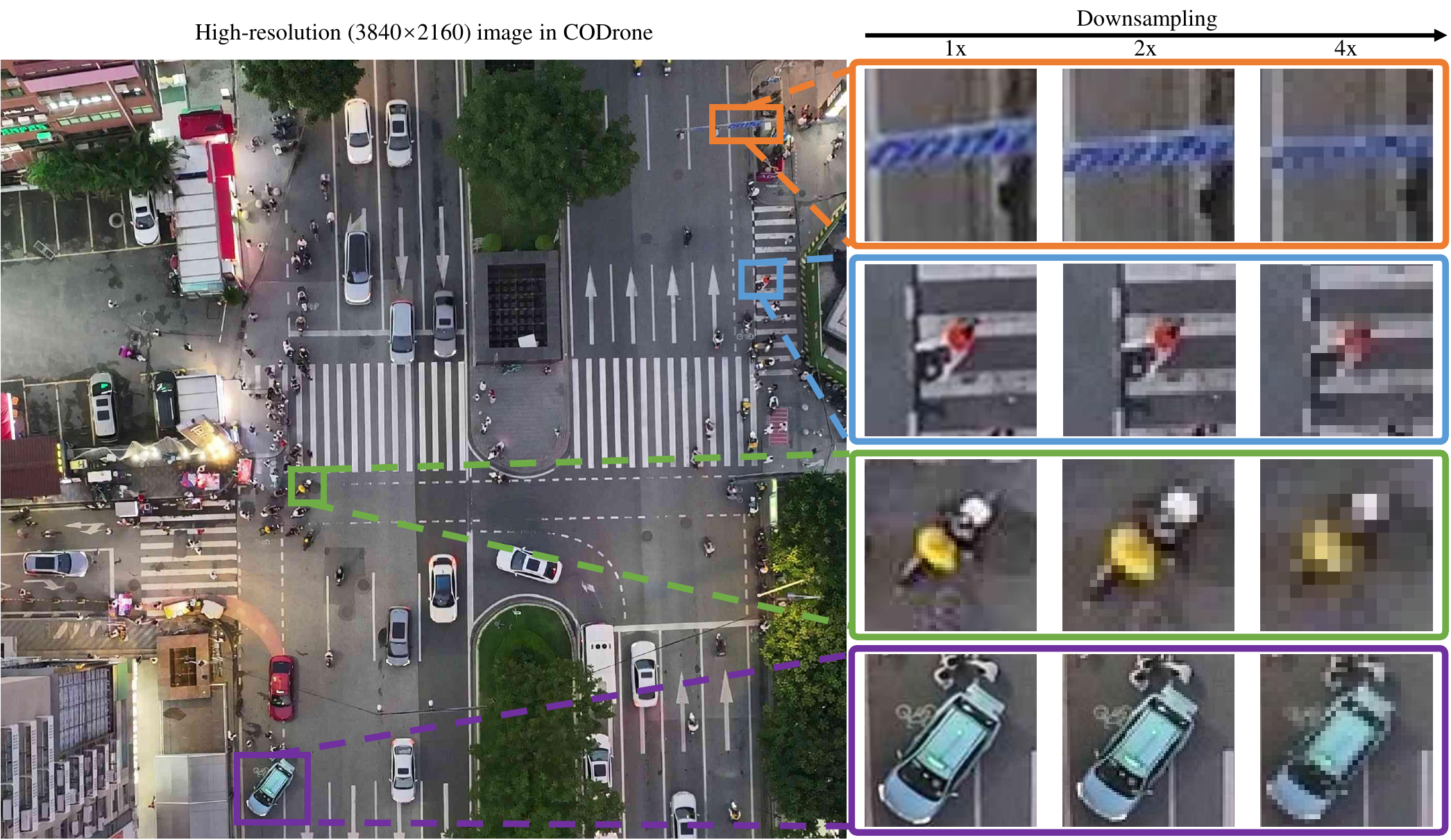}
    \caption{Visualization of object features at different resolutions.
    The left image shows a cropped region from an original CODrone image.
    The right side illustrates the appearance of the same objects under different downsampling scales.
    “1×” denotes the original resolution, while “2×” and “4×” correspond to 2× and 4× downsampling, respectively.}
    \label{fig:res}
\end{figure*}
\subsubsection{High resolution brings more high-quality information}
As with many remote sensing perception tasks, advancements in sensor technology have significantly increased the density and richness of captured information, introducing new challenges for downstream algorithms.
In the context of UAV-based oriented object detection, high-resolution imagery enables more accurate and detailed annotation, particularly for categories that were previously difficult to label precisely due to blurred or indistinct features.
As illustrated in Fig. \ref{fig:res}, small objects such as pedestrians and bicycles exhibit significant differences in feature representation across different resolutions.
At low resolutions, these objects often appear as indistinguishable noise or blur, making precise oriented annotation infeasible.
In contrast, high-resolution imagery preserves clear structural details, allowing for reliable and fine-grained OBB (oriented bounding box) annotation.

For larger objects like vehicles, high-resolution input further enhances annotation precision by providing sharper edge information, which facilitates more accurate estimation of object orientation and shape.
These observations highlight the critical role of image resolution in improving both the quality of annotations and the effectiveness of detection algorithms, especially for small or densely packed objects.

\subsubsection{Multi-altitude and multi-angle captures for broad flight scenario adaptation}
To accommodate a broader range of UAV flight conditions, CODrone was designed with specific requirements for both flight altitudes and camera angles, allowing effective perception under various operational scenarios.
The lowest flight altitude was set to 30 meters, which we consider a relatively safe minimum height in urban environments.
At this altitude, the UAV is unlikely to be affected by obstacles such as utility poles, pedestrian overpasses, or tall vegetation.

\begin{figure}
    \centering
    \includegraphics[width=1\linewidth]{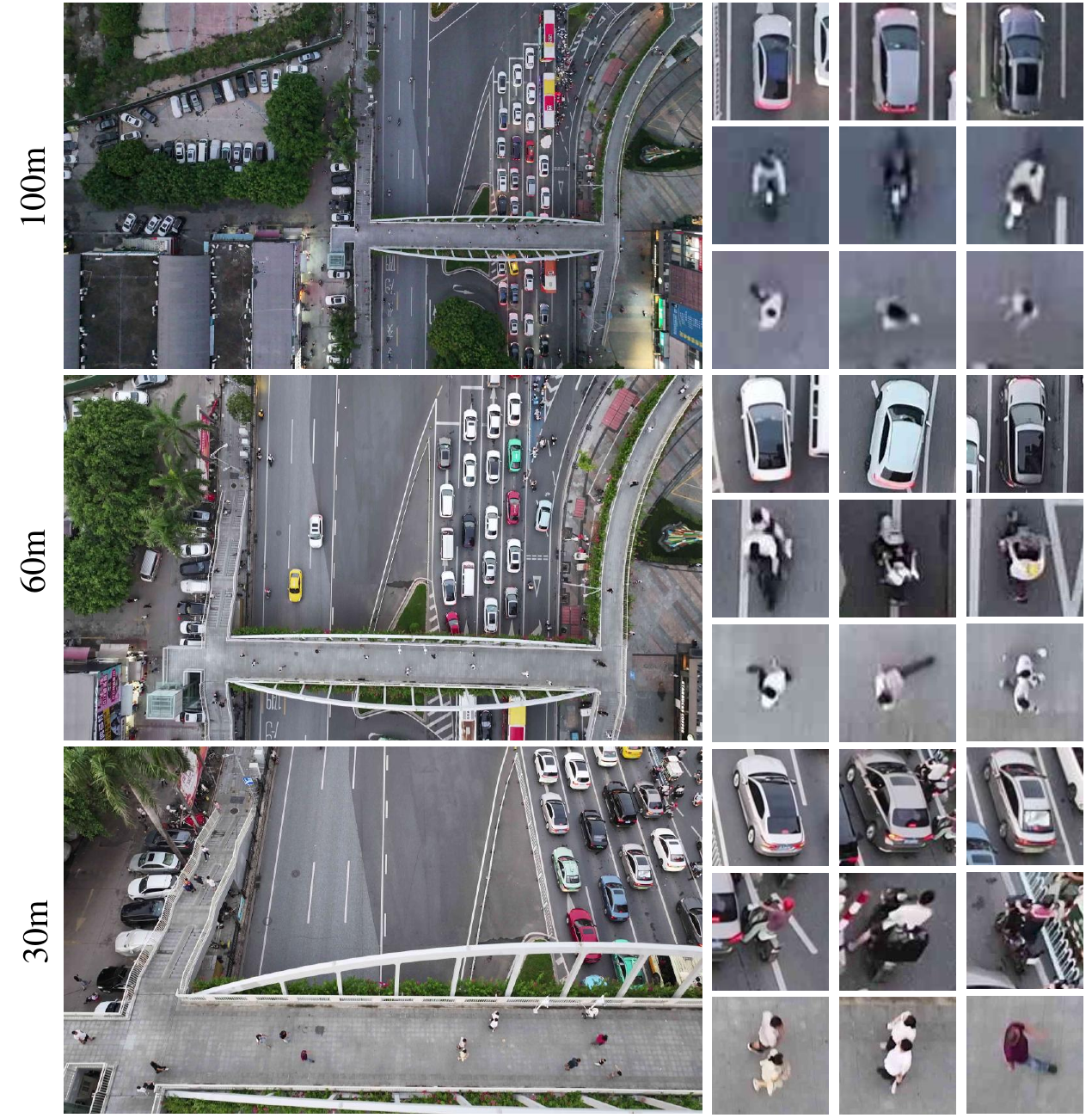}
    \caption{ Visualization of the altitude variation on object features. 
    Using car, motorcycle, and pedestrian as examples, we observe significant differences in object appearance across different altitude levels. 
    Notably, intra-class feature variation increases with altitude shifts, resulting in larger intra-class distances.}%
    \label{fig:height}
\end{figure}

The highest altitude was set to 100 meters, taking into account the typical altitude limit of small UAVs, which is approximately 120 meters in most regulatory frameworks.
A 20-meter safety margin was reserved to ensure compliance and operational flexibility.
For intermediate altitude, we selected 60 meters as a balanced choice that provides both a sufficient field of view and adequate object detail.
As illustrated in Fig. \ref{fig:height}, the same scene captured from different altitudes shows substantial visual variation.This significantly increases intra-class appearance diversity, presenting a greater challenge for OOD algorithms to generalize across altitude changes.

In terms of camera angle selection, we adopt two representative viewpoints commonly used in UAV perception.
The first is a 90-degree nadir view, i.e., a strictly top-down perspective, which is consistent with the setup of most existing remote sensing OOD datasets.
The second is a 30-degree oblique view, which better reflects the operational characteristics of real-world UAV deployments.
We note that in actual UAV flights, minor deviations in flight posture often introduce natural variations in the camera's orientation. 
As a result, viewpoints within the intermediate range between 30° and 90° tend to yield image appearances that are visually similar to either of these two endpoints.
Therefore, instead of densely sampling intermediate angles, we select 30° and 90° as two sufficiently distinct viewpoints that effectively capture the diversity of real-world UAV perspectives.

\begin{figure}
    \centering
    \includegraphics[width=1\linewidth]{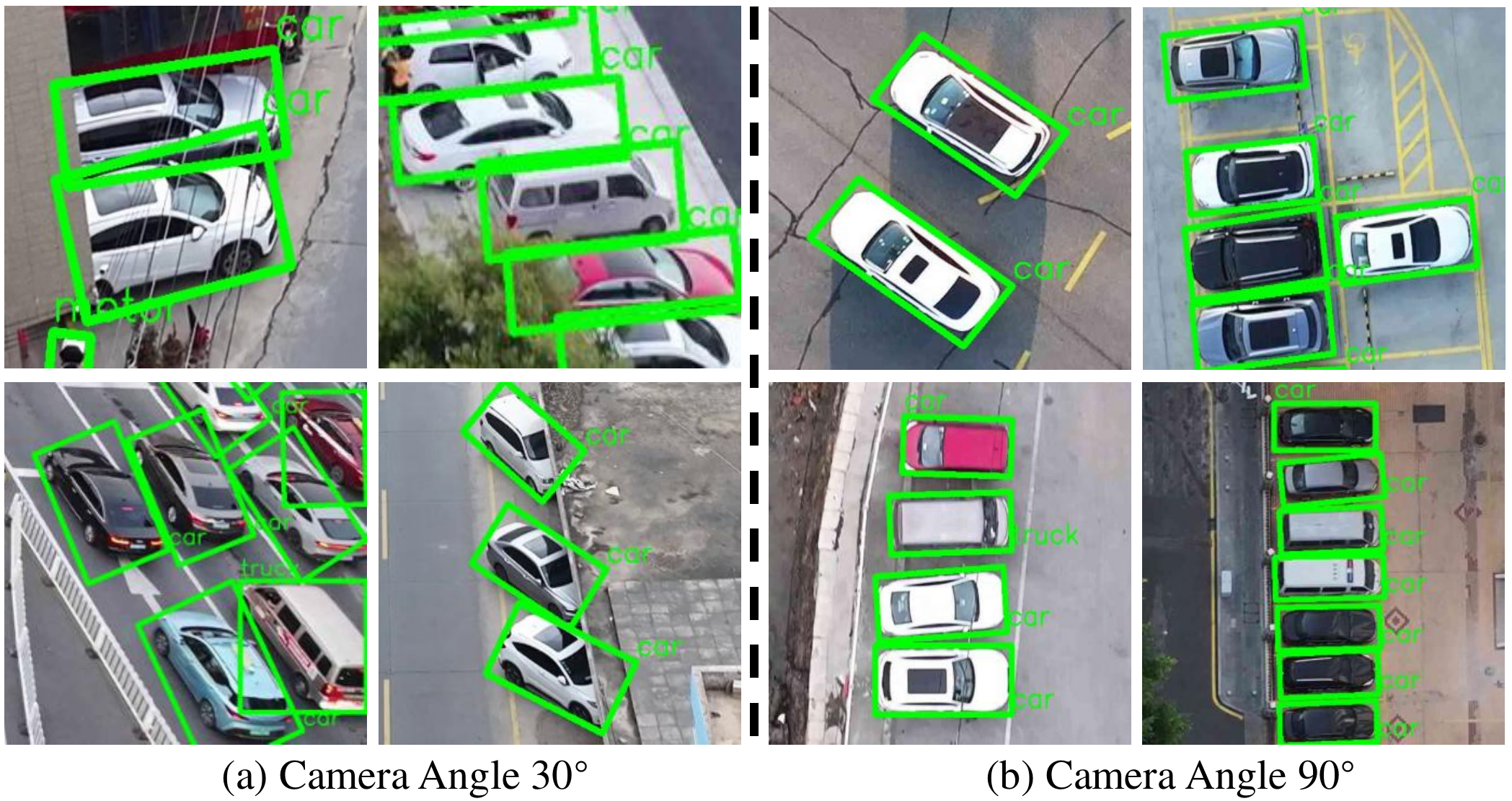}
    \caption{Visualization of oriented bounding box annotations under two different camera angles. Taking vehicles as an example, (a) shows annotations captured at a 30-degree oblique angle, while (b) shows annotations from a top-down 90-degree view.}
    \label{fig:angle}
\end{figure}

Importantly, oriented object detection under non-vertical perspectives introduces additional complexity. 
As shown in Fig. \ref{fig:angle}, under a 30-degree viewing angle, the correspondence between object orientation and feature representation becomes more ambiguous. 
In contrast, the nadir view provides a simplified top-down appearance, where the object’s rotation is more easily defined. 
However, the oblique view introduces richer visual cues, such as side profiles and foreshortening effects, which require the detection model to understand the relationship between features and rotation angles from fundamentally different perspectives.

The combination of three flight altitudes and two camera angles in CODrone results in six distinct viewpoint configurations. 
As shown in Table \ref{h}, the dataset is approximately evenly distributed across these six combinations, ensuring balanced representation for each flight condition.
To facilitate more targeted research under specific UAV flight postures, the corresponding altitude and angle information is encoded in the suffix of each image filename.
This enables researchers to easily filter or analyze subsets of the data based on specific viewpoint configurations.

\begin{table}
  \centering
  \caption{Distribution of images at different flight altitude and camera angles}
    \begin{tabular}{c|c|cc}
    \toprule
    Flight altitude & Camera angle & Number of pictures & Ratio (\%) \\
    \midrule
    \multirow{2}*[-0.2ex]{30m} & 30°   & 1,865 & 18.64 \\
          & 90°   & 1,276 & 12.75 \\
    \midrule
    \multirow{2}*[-0.2ex]{60m} & 30°   & 2,047 & 20.46 \\
          & 90°   & 2,269 & 22.68 \\
    \midrule
    \multirow{2}*[-0.2ex]{90m} & 30°   & 1,048 & 10.48 \\
          & 90°   & 1,499 & 14.98 \\
    \bottomrule
    \end{tabular}%
  \label{h}%
\end{table}%

\subsubsection{More diverse scenes, broader application potential}
CODrone covers a wide range of environments, from urban areas and rural towns to ports and industrial zones, encompassing most scene types encountered in real-world UAV-based urban applications.
To the best of our knowledge, it is the only publicly available UAV OOD dataset that spans such diverse scene categories.
This broad coverage greatly enriches the variety of object classes included in the dataset. 
For example, categories like ship and boat, which typically appear in coastal environments, are represented with multiple annotated instances. 

Additionally, considering UAV operation under low-light conditions, CODrone includes imagery captured at night, with corresponding scene distributions shown in Table \ref{l}.
It is important to note that although we categorize lighting conditions broadly as daytime and nighttime, the actual images exhibit a wide range of illumination variations due to differences in weather, seasons, and time of day. 
This further enhances the illumination diversity of the dataset.

\begin{table}
  \centering
  \caption{Distribution under different light conditions}
    \begin{tabular}{c|cc}
    \toprule
    Scene & Daytime   & Nighttime \\
    \midrule
    Number of pictures & 6,121  & 3,883 \\
    Ratio (\%) & 0.61 & 0.39 \\
    \bottomrule
    \end{tabular}%
  \label{l}%
\end{table}%

\section{Benchmark Construction}
\subsection{Evaluation metrics}

To evaluate algorithm performance on CODrone, we adopt two widely used metrics in oriented object detection tasks: AP50 and AP75.
These metrics are based on the standard definition of Average Precision (AP) at different IoU thresholds, tailored to assess the quality of both detection and localization.
The Average Precision (AP) is calculated as the area under the precision-recall curve:
\begin{align}
  R &= \frac{TP}{TP + FN}\\
P &= \frac{TP}{TP + FP}\\
AP &= \int_{0}^{1} P(R) \, dR
\end{align}
where $TP$, $FP$, and $FN$ denote the number of true positives, false positives, and false negatives, respectively.
The recall $R$ measures the proportion of ground-truth instances that are correctly detected by the model. 
$P$ is precision, $P(R)$ is the precision as a function of recall.

In the context of oriented object detection, AP50 evaluates detection performance at an Intersection over Union (IoU) threshold of 0.5, while AP75 uses a stricter threshold of 0.75. The IoU between predicted oriented bounding box $B_p$ and ground truth box $B_{gt}$ is defined as:
\begin{equation}
IoU=\frac{Area(B_p\cap B_{gt})}{Area(B_p\cup B_{gt})}    
\end{equation}

Extensive studies and empirical results have shown that AP75 provides greater sensitivity in evaluating rotation-aware detection performance\cite{DBLP:conf/nips/YangYYMWTY21,DBLP:journals/corr/abs-2303-04989}.
This is particularly critical in OOD tasks, where box orientation significantly influences task performance, especially for elongated objects (e.g., ships, bridges, vehicles) whose aspect ratios amplify small angular errors.

A small misalignment in angle might still result in a relatively high IoU at 0.5, but it may drop below 0.75, especially when the predicted box is mis-rotated yet still overlaps substantially. 
Therefore, AP75 penalizes angular deviations more strictly, making it a better indicator of precise localization and orientation accuracy.

\subsection{Selected algorithms}
To establish a benchmark on the CODrone dataset, we select 22 representative or SOTA methods that are either widely used in the community or have achieved leading performance under different settings and task formulations.
In this section, we provide a brief overview of each selected method, highlighting their core characteristics and design principles. 
This serves as a foundation for the subsequent performance analysis, where we examine how different algorithms respond to the unique challenges posed by CODrone.

\begin{itemize}
  \item \textbf{Rotated Faster R-CNN} (TPAMI 2017)\cite{DBLP:journals/pami/RenHG017}: An extension of Faster R-CNN to support oriented bounding boxes, enabling detection of oriented objects in remote sensing images.
  
  \item \textbf{Rotated RetinaNet} (ICCV 2017)\cite{DBLP:conf/iccv/LinGGHD17}: Adapts RetinaNet into a single-stage detector capable of handling oriented bounding boxes, improving its ability to detect oriented objects.
  
  \item \textbf{Rotated RepPoints} (ICCV 2019)\cite{DBLP:conf/iccv/YangLHWL19}: A keypoint-based detection framework that enhances representation and localization of oriented objects.
  
  \item \textbf{RoI Transformer} (CVPR 2019)\cite{DBLP:conf/cvpr/DingXLXL19}: Proposes the supervised rotated RoI learner and rotated position-sensitive RoI alignment modules, which effectively alleviate the misalignment between horizontal RoIs and oriented objects, as well as the associated complexity issues.
  
  \item \textbf{CSL} (ECCV 2020)\cite{DBLP:conf/eccv/0005Y20}: Proposes a circular smooth label encoding scheme to alleviate boundary discontinuities in angle regression.
  
  \item \textbf{Rotated ATSS} (CVPR 2020)\cite{DBLP:conf/cvpr/ZhangCYLL20}: Extends the Adaptive Training Sample Selection mechanism to oriented object detection, improving the adaptability of sample selection.
  
  \item \textbf{Gliding Vertex} (TPAMI 2020)\cite{DBLP:journals/pami/XuFWWCXB21}: Introduces a vertex sliding mechanism to avoid the discontinuity problem in angle regression and the square-like box problem, thereby reducing the learning difficulty for the model.
  
  \item \textbf{Oriented R-CNN} (ICCV 2021)\cite{DBLP:conf/iccv/Xie0WYH21}: Proposes Oriented RPN by introducing two additional regression parameters in the RPN branch, resulting in a six-parameter representation for oriented objects. This approach avoids the order ambiguity in corner-based regression and provides constraints for oriented bounding box regression.
  
  \item \textbf{GWD} (ICML 2021)\cite{DBLP:conf/icml/0005YMWZ021}:Uses the Gaussian Wasserstein Distance (GWD) to measure the distance between rotated bounding boxes, replacing the traditional IoU loss with a differentiable GWD-based loss.
  This approach addresses the discontinuity problem in angle regression and the square-like box issue, while being agnostic to the specific parameterization of bounding boxes. 
  
  \item \textbf{KLD} (NeurIPS 2021)\cite{DBLP:conf/nips/YangYYMWTY21}: Converts rotated bounding boxes into 2D Gaussian distributions and computes the Kullback-Leibler Divergence (KLD) between them as the regression loss. KLD and its derivatives adaptively adjust the gradient weights of the angle parameters according to the aspect ratio of the objects.

  \item \textbf{S2ANet} (TGRS 2021)\cite{DBLP:journals/tgrs/HanDLX22}: 
  Proposes the feature alignment module, which generates high-quality anchors through the Anchor Refinement Network and employs aligned convolution to adaptively align convolutional features with the anchor boxes. Additionally, the oriented detection module utilizes active rotating filters to encode orientation information, mitigating the inconsistency between classification scores and localization accuracy.
  
  \item \textbf{R3Det} (AAAI 2021)\cite{DBLP:conf/aaai/YangYFH21}: 
  Proposes to first generate proposals rapidly using horizontal anchors, and then refine them with rotated anchors through a feature refinement module to enable fast and accurate oriented object detection in dense scenes.
  
  \item \textbf{ReDet} (CVPR 2021)\cite{DBLP:conf/cvpr/HanD0X21}: Employs rotation-invariant feature extractors to enhance robustness in oriented object detection.
  
  \item \textbf{Oriented RepPoints} (CVPR 2022)\cite{DBLP:conf/cvpr/LiCHZ22}: 
  Proposes an effective adaptive point learning method that captures the geometric information of arbitrarily oriented instances through adaptive point representations.
  A quality assessment and sample assignment scheme are introduced for adaptive points, designed to better capture non-axis-aligned features from neighboring objects or background noise.
  
  \item \textbf{KFIoU} (ICLR 2023)\cite{DBLP:conf/iclr/00050ZYWY0023}: 
  Proposes a loss function based on Gaussian modeling and Gaussian product to effectively approximate SkewIoU. It consists of two components: (1) a scale-insensitive center point loss to quickly reduce the distance between the centers of two bounding boxes, (2) a distance-independent term that leverages the product of Gaussian distributions to mimic the mechanism of SkewIoU.
  
  \item \textbf{H2RBox} (ICLR 2023)\cite{DBLP:conf/iclr/0005ZL0WY23}: 
  Achieves oriented object detection using only horizontal bounding box annotations under a weakly supervised training paradigm. 
  The core approach combines weakly supervised learning and self-supervised learning by enforcing consistency between two different views to predict object features.
  
  \item \textbf{DCFL} (CVPR 2023)\cite{DBLP:conf/cvpr/XuDWYYYX23}: 
  Combines the outputs of deep neural networks and dynamically redefines priors, label assignment, and ground truth representation to alleviate the incorrect matching problem. In addition, DCFL adopts a coarse-to-fine strategy to progressively and dynamically assign labels during training.
  
  \item \textbf{PSC} (CVPR 2023)\cite{DBLP:conf/cvpr/0010D23}: 
  Provides a unified framework to address various periodic ambiguities caused by rotational symmetry, such as boundary discontinuities and square-like issues, by mapping different periods of rotational periodicity to phases of different frequencies.
  
  \item \textbf{LSKNet} (ICCV 2023)\cite{DBLP:conf/iccv/LiHZCYL23}: 
  Explores large and selective kernel mechanisms in remote sensing object detection, enabling the dynamic adjustment of large spatial receptive fields to better model the varying contextual scales of objects in remote sensing scenarios.
  
  \item \textbf{H2RBox-v2} (NeurIPS 2023)\cite{DBLP:conf/nips/YuYL0DY23}: 
  Builds on H2RBox\cite{DBLP:conf/iclr/0005ZL0WY23} by introducing a self-supervised branch with a novel symmetry-aware learning paradigm, which operates independently of the weakly supervised branch and learns object orientation directly from images based on object symmetry.
  
  
  
  \item \textbf{OrientedFormer} (TGRS 2024)\cite{DBLP:journals/tgrs/ZhaoDZZDYE24}: 
  Proposes Wasserstein self-attention to introduce geometric relations and facilitate the interaction between content and positional queries by utilizing Gaussian Wasserstein distance scores. Oriented cross-attention is proposed to align values with positional queries by rotating sampling points around the positional queries according to their angles.
  
  \item \textbf{DeCoupleNet} (TGRS 2024)\cite{DBLP:journals/tgrs/LuCSTL24}: 
  Introduces the feature integration downsampling module to preserve small object features during downsampling and the multibranch feature decoupling module to enhance the feature representation of small and multiscale objects through a novel decoupling approach.
  
\end{itemize}

Meanwhile, this serves as a concise summary of current methods in oriented object detection, intended to support further developments by the research community.

\begin{table*}[htbp]
  \centering
  \caption{Performance Comparison of Selected Oriented Object Detection Methods on the CODrone}
    \begin{tabular}{ccc|cc|cc}
    \toprule
    Model & Source & Backbone & AP50  & Rank  & AP75  & Rank \\
    \midrule
    Rotated Faster R-CNN\cite{DBLP:journals/pami/RenHG017} & TPAMI'2017 & ResNet 50 & 40.40 & 14    & 15.23 & 18 \\
    Rotated RetinaNet\cite{DBLP:conf/iccv/LinGGHD17} & ICCV'2017 & ResNet 50 & 40.62 & 12    & 18.26 & 8 \\
    Rotated RepPoints\cite{DBLP:conf/iccv/YangLHWL19} & ICCV'2019 & ResNet 50 & 31.25 & 22    & 13.21 & 19 \\
    RoI Transformer\cite{DBLP:conf/cvpr/DingXLXL19} & CVPR'2019 & ResNet 50 & 43.03 & 4     & 19.98 & 3 \\
    CSL\cite{DBLP:conf/eccv/0005Y20}   & ECCV'2020 & ResNet 50 & 40.05 & 16    & 17.28 & 14 \\
    Rotated ATSS\cite{DBLP:conf/cvpr/ZhangCYLL20} & CVPR'2020 & ResNet 50 & 42.85 & 5     & 19.19 & 5 \\
    Gliding Vertex\cite{DBLP:journals/pami/XuFWWCXB21} & TPAMI'2020 & ResNet 50 & 41.71 & 8     & 15.43 & 17 \\
    Oriented R-CNN\cite{DBLP:conf/iccv/Xie0WYH21} & ICCV'2021 & ResNet 50 & 42.33 & 7     & 18.99 & 6 \\
    GWD\cite{DBLP:conf/icml/0005YMWZ021}   & ICML'2021 & ResNet 50 & 40.51 & 13    & 18.04 & 10 \\
    KLD\cite{DBLP:conf/nips/YangYYMWTY21}   & NeurIPS'2021 & ResNet 50 & 40.98 & 10    & 17.73 & 12 \\
    S2ANet\cite{DBLP:journals/tgrs/HanDLX22}  & TGRS'2021 & ResNet 50 & 37.61 & 20    & 10.61 & 21 \\
    R3Det\cite{DBLP:conf/aaai/YangYFH21}  & AAAI'2021 & ResNet 50 & 39.88 & 18    & 16.28 & 16 \\
    ReDet\cite{DBLP:conf/cvpr/HanD0X21}  & CVPR'2021 & ReResNet 50 & 44.73 & 2     & 20.17 & 2 \\
    Oriented RepPoints\cite{DBLP:conf/cvpr/LiCHZ22}  & CVPR'2022 & ResNet 50 & 44.59 & 3     & 19.62 & 4 \\
    KFIoU\cite{DBLP:conf/iclr/00050ZYWY0023}  & ICLR'2023 & ResNet 50 & 40.63 & 11    & 18.13 & 9 \\
    H2RBox\cite{DBLP:conf/iclr/0005ZL0WY23} & ICLR'2023 & ResNet 50 & 36.22 & 21    & 10.29 & 22 \\
    DCFL\cite{DBLP:conf/cvpr/XuDWYYYX23}  & CVPR'2023 & ResNet 50 & 42.77 & 6     & 17.68 & 13 \\
    PSC\cite{DBLP:conf/cvpr/0010D23}   & CVPR'2023 & ResNet 50 & 40.13 & 15    & 18.93 & 7 \\
    LSKNet\cite{DBLP:conf/iccv/LiHZCYL23} & ICCV'2023 & LSKNet-S & 46.92 & 1     & 21.15 & 1 \\
    H2RBox-v2\cite{DBLP:conf/nips/YuYL0DY23} & NeurIPS'2023 & ResNet 50 & 39.07 & 19    & 12.21 & 20 \\
    OrientFormer\cite{DBLP:journals/tgrs/ZhaoDZZDYE24}  & TGRS'2024  & ResNet 50 & 41.02 & 9     & 18.00 & 11 \\
    DeCoupleNet\cite{DBLP:journals/tgrs/LuCSTL24}  & TGRS'2024 & DeCoupleNet D0 & 39.90 & 17    & 16.70 & 15 \\
    \bottomrule
    \end{tabular}%
  \label{tab:E1}%
\end{table*}%

\subsection{Implementation details}
All models are trained on a single NVIDIA GTX 3090 GPU.
To ensure a fair comparison and to remain consistent with the original implementation of each method, we adopt the same batch size, optimizer, and learning rate settings as reported in the respective papers or official codebases.
Each algorithm is trained for 30 epochs, and we report the best-performing result to ensure that the model is sufficiently trained.
For all methods, we used a momentum of 0.9 and a weight decay of 0.00005, following commonly adopted optimization practices in the field.
Due to the high image resolution and the small size of the objects, we follow the DOTA dataset's partitioning strategy by splitting the images into smaller patches for training.

\subsection{Algorithm performance analyze}
As shown in Table~\ref{tab:E1}, the performance of 22 selected oriented object detection methods exhibits significant variation in both the standard AP50 metric and the more stringent AP75 metric.
These differences reflect the distinct characteristics and technical strategies of each method when applied to the challenging task of detecting arbitrarily oriented objects.
The evaluation, conducted on the proposed CODrone dataset, further reveals the complexity of this benchmark, which features diverse scenes, densely packed small objects, and a wide range of object orientations.

Among the evaluated methods, LSKNet~\cite{DBLP:conf/iccv/LiHZCYL23} and ReDet~\cite{DBLP:conf/cvpr/HanD0X21} demonstrate superior performance. 
LSKNet achieves the highest AP50 (46.92) and ranks second in AP75 (21.15), highlighting its effectiveness in handling diverse scenarios typical in UAV imagery. 
Specifically, the use of large-kernel convolutional sequences combined with a spatial selection mechanism allows LSKNet to dynamically adjust its receptive field, making it suited for varying backgrounds and viewpoint shifts. 
Such dynamic modeling of contextual information is crucial for OOD in CODrone, where objects are observed across a wide range of scales, densities, and environmental conditions.
In contrast, ReDet achieves the second-highest AP50 (44.73) and the highest AP75 (20.17), demonstrating outstanding robustness under stricter localization requirements. 
By modeling orientation variations starting from the backbone network, ReDet enhances its ability to capture the geometric diversity of objects, especially under oblique viewing angles common in UAV scenarios. 
This early incorporation of rotation-awareness throughout the feature hierarchy enables more precise detection and boundary alignment across complex, non-vertically aligned targets present in CODrone.

Several other methods, including RoI Transformer~\cite{DBLP:conf/cvpr/DingXLXL19}, Oriented RepPoints~\cite{DBLP:conf/cvpr/LiCHZ22}, and Rotated ATSS~\cite{DBLP:conf/cvpr/ZhangCYLL20}, also perform competitively.
RoI Transformer achieves an AP50 of 43.03 and an AP75 of 19.98, demonstrating the effectiveness of learning affine transformations for correcting rotational misalignment.
Oriented RepPoints, with its rotation-sensitive point-based representation, ranks third in AP50 (44.59) and excels in localization accuracy.
Rotated ATSS leverages an adaptive sample selection strategy that contributes to stable performance across object scales and orientations, ranking fifth in both metrics.

\begin{table*}[htbp]
  \centering
  \caption{Performance comparison of selected methods under different viewpoint combinations on the CODrone dataset}
    \begin{tabular}{c|cc|cc|cc|cc|cc|cc}
    \toprule
    \multirow{2}*[-0.2ex]{Model} & \multicolumn{2}{c|}{30m30°} & \multicolumn{2}{c|}{30m90°} & \multicolumn{2}{c|}{60m30°} & \multicolumn{2}{c|}{60m90°} & \multicolumn{2}{c|}{100m30°} & \multicolumn{2}{c}{100m90°} \\
          & AP50  & AP75  & AP50  & AP75  & AP50  & AP75  & AP50  & AP75  & AP50  & AP75  & AP50  & AP75 \\
    \midrule
    Rotated Faster R-CNN & 39.13  & 14.32  & 39.77  & 14.86  & 39.58  & 14.55  & 45.05  & 19.51  & 32.72  & 11.13  & 44.56  & 20.77  \\
    Rotated RetinaNet & 38.69  & 15.97  & 39.54  & 19.92  & 39.09  & 17.31  & 45.37  & 23.68  & 31.59  & 13.03  & 44.13  & 24.47  \\
    Rotated RepPoints & 30.12  & 12.26  & 29.78  & 12.68  & 29.89  & 11.77  & 36.41  & 16.17  & 24.65  & 9.54  & 33.01  & 15.38  \\
    RoI Transformer & 40.62  & 18.46  & 44.73  & 22.67  & 41.12  & 17.73  & 48.67  & 25.26  & 34.67  & 13.46  & 47.12  & 26.53  \\
    CSL   & 38.99  & 16.48  & 41.28  & 18.95  & 38.09  & 16.58  & 45.28  & 21.71  & 31.52  & 12.21  & 43.59  & 23.15  \\
    Rotated ATSS & 40.19  & 17.58  & 40.18  & 18.77  & 40.42  & 17.78  & 48.39  & 24.71  & 34.67  & 14.05  & 48.98  & 28.46  \\
    Gliding Vertex & 39.00  & 13.77  & 40.46  & 16.73  & 40.50  & 15.67  & 46.76  & 19.59  & 34.18  & 11.38  & 45.95  & 23.06  \\
    Oriented R-CNN & 39.18  & 17.53  & 44.03  & 18.30  & 40.85  & 18.56  & 48.57  & 24.38  & 35.18  & 12.95  & 46.99  & 26.26  \\
    GWD   & 37.99  & 17.32  & 41.57  & 23.01  & 38.41  & 17.91  & 46.23  & 24.49  & 32.71  & 13.98  & 43.16  & 24.33  \\
    KLD   & 38.70  & 16.78  & 40.96  & 18.65  & 39.33  & 17.34  & 45.46  & 24.19  & 32.91  & 13.19  & 42.70  & 22.48  \\
    S2ANet  & 35.95  & 10.03  & 41.00  & 16.69  & 36.38  & 10.52  & 40.06  & 11.99  & 31.42  & 8.17  & 40.19  & 13.27  \\
    R3Det  & 37.93  & 14.26  & 39.20  & 15.92  & 38.25  & 14.40  & 45.92  & 20.19  & 32.41  & 11.82  & 45.13  & 24.29  \\
    ReDet  & 41.65  & 19.19  & 49.99  & 24.38  & 41.49  & 18.61  & 48.77  & 26.19  & 35.56  & 14.01  & 49.12  & 27.85  \\
    Oriented RepPoints  & 42.92  & 18.97  & 44.36  & 22.69  & 42.92  & 17.89  & 48.42  & 24.33  & 37.75  & 15.41  & 47.49  & 26.66  \\
    KFIoU  & 38.26  & 16.55  & 39.81  & 18.70  & 38.62  & 17.49  & 45.23  & 21.79  & 31.82  & 13.15  & 43.62  & 22.58  \\
    H2RBox & 37.90  & 10.80  & 50.20  & 14.90  & 34.00  & 8.40  & 40.10  & 13.80  & 28.30  & 6.20  & 37.40  & 11.30  \\
    DCFL  & 40.77  & 16.37  & 44.43  & 20.76  & 41.23  & 16.77  & 46.86  & 22.98  & 35.21  & 13.61  & 46.72  & 25.15  \\
    PSC   & 39.70  & 17.50  & 44.70  & 24.00  & 37.10  & 16.70  & 45.80  & 25.40  & 28.90  & 11.90  & 42.70  & 24.40  \\
    LSKNet & 43.87  & 19.75  & 55.71  & 32.38  & 43.86  & 19.34  & 51.82  & 27.17  & 38.63  & 16.45  & 51.31  & 28.23  \\
    H2RBox-v2 & 38.90  & 13.10  & 48.60  & 14.30  & 36.00  & 9.90  & 46.10  & 16.10  & 29.40  & 7.70  & 43.80  & 15.00  \\
    OrientFormer  & 40.61  & 18.02  & 47.03  & 23.40  & 39.70  & 18.10  & 46.03  & 24.70  & 32.10  & 12.20  & 43.40  & 27.10  \\
    DeCoupleNet  & 38.60 & 17.20 & 45.10 & 22.30 & 37.10 & 16.20 & 44.50 & 23.90 & 30.60 & 11.40 & 42.90 & 23.60 \\
    \bottomrule
    \end{tabular}%
  \label{tab:e2}%
\end{table*}%

Classical architectures such as Rotated Faster R-CNN~\cite{DBLP:journals/pami/RenHG017}, Rotated RetinaNet~\cite{DBLP:conf/iccv/LinGGHD17}, and Gliding Vertex~\cite{DBLP:journals/pami/XuFWWCXB21} maintain relatively robust performance despite their earlier designs.
For instance, Rotated RetinaNet ranks 12th in AP50 (40.62) but improves to 8th in AP75 (18.26), suggesting that focal loss-based single-stage detectors retain competitive precision at higher thresholds.

Recent methods based on distance-based regression loss, such as GWD~\cite{DBLP:conf/icml/0005YMWZ021} and KLD~\cite{DBLP:conf/nips/YangYYMWTY21}, show stable results in AP50 (ranking 13th and 10th, respectively), indicating their benefit in improving regression stability.
Their relatively lower AP75 scores suggest that such approaches may still lack fine-grained boundary localization capabilities under strict evaluation conditions, which are crucial in datasets like CODrone where accurate angle and boundary regression is critical.
Transformer-based models, such as OrientFormer~\cite{DBLP:journals/tgrs/ZhaoDZZDYE24}, demonstrate promising potential, ranking 9th in AP50.
However, its performance at AP75 (11th) reveals limitations in precise boundary modeling, potentially due to insufficient integration of local spatial cues. 

At the lower end of the spectrum, methods such as Rotated RepPoints, H2RBox, and H2RBox-v2 perform poorly in both AP50 and AP75 (below rank 20), indicating that insufficient geometric modeling or overly complex decoding processes can hinder accurate localization.
In CODrone, where objects appear at arbitrary orientations and scales across diverse scenes, the failure to capture such spatial variation directly impacts detection performance.

In summary, the experimental results highlight that recent advancements in oriented object detection benefit substantially from robust feature transformation, adaptive anchor design, and attention-based mechanisms.
For more challenging benchmark, CODrone, further improvements are needed in spatial feature alignment, integration of local-global context, and enhancing rotation sensitivity within different frameworks to achieve high-precision detection.

\subsection{Effects of flight altitude and imaging angle}
Table~\ref{tab:e2} presents a comprehensive performance comparison of representative oriented object detection (OOD) methods under various viewpoint combinations on the proposed CODrone dataset. 
The CODrone dataset is characterized by its extensive coverage of different flight altitudes (30m, 60m, 100m) and camera tilt angles (30°, 90°), thereby posing unique challenges for UAV-based OOD.

Several important trends can be observed. 
First, across all models, performance generally deteriorates with increasing altitude, particularly in terms of AP75 scores, reflecting the difficulty of maintaining precise localization at greater distances.
For instance, when comparing results between 30m30° and 100m30° settings, models such as Rotated Faster R-CNN and Rotated RetinaNet exhibit notable degradations in both AP50 and AP75, underscoring the challenges in detecting smaller and less distinguishable objects from higher viewpoints.

Second, variations in camera tilt angles introduce additional complexities. 
Detection models consistently achieve better performance under 90° camera tilt settings compared to 30°, suggesting that nadir-like views (i.e., vertical shots) suffer less from occlusions and perspective distortions than oblique views. 
Methods such as RoI Transformer, ReDet, and LSKNet exhibit substantial performance gaps between 30° and 90° configurations, particularly at higher altitudes (e.g., 100m), highlighting the difficulty of modeling objects under severe perspective deformation.

Third, although state-of-the-art approaches such as LSKNet, ReDet, and Oriented RepPoints achieve leading results, their AP75 values remain substantially lower than their AP50 scores, especially under challenging settings like 60m30° and 100m30°. 
This disparity suggests that precise bounding box regression remains a major challenge under CODrone’s real-world UAV flight conditions, where object sizes, background clutter, and wide-angle distortions vary significantly.

The experimental results reveal that CODrone presents substantial challenges to existing OOD algorithms, particularly in terms of robustness across wide altitude-angle variations, sensitivity to perspective effects, and demands for fine-grained localization accuracy. 
These findings indicate that further methodological innovations are required to bridge the performance gap and enhance detection robustness under realistic UAV deployment scenarios.

\section{Conclusion}
In this work, we introduced CODrone, a comprehensive and high-resolution dataset specifically designed for oriented object detection (OOD) in UAV scenarios. 
Recognizing the limitations of existing datasets—such as low resolution, limited object diversity, fixed altitudes, and restricted camera perspectives—we systematically addressed these gaps by collecting over 10,000 images across multiple cities, covering diverse urban and industrial environments under varying lighting conditions, altitudes, and camera angles.
Compared to existing benchmarks, CODrone captures the complexity and variability inherent to real-world UAV operations, such as significant changes in object scale, diverse viewpoint-induced distortions, and cluttered background conditions. 
Through extensive experiments involving 22 representative and state-of-the-art methods, we systematically evaluated the challenges posed by CODrone, revealing significant performance gaps under varying altitudes and oblique viewing angles. 
These results highlight critical limitations in current OOD algorithms, particularly regarding their robustness to spatial transformations and fine-grained localization under realistic UAV imaging conditions.
We hope that CODrone will serve as a valuable resource for the community, promoting advancements in both academic research and practical UAV applications in fields such as logistics, urban management, disaster response, and autonomous inspection.
The CODrone dataset is publicly available at \url{https://github.com/AHideoKuzeA/CODrone-A-Comprehensive-Oriented-Object-Detection-benchmark-for-UAV}, and we encourage future work to build upon the challenges and insights identified through this benchmark.




\bibliographystyle{IEEEtran}
\bibliography{example_paper}

\begin{thebibliography}{10}
\providecommand{\url}[1]{#1}
\csname url@samestyle\endcsname
\providecommand{\newblock}{\relax}
\providecommand{\bibinfo}[2]{#2}
\providecommand{\BIBentrySTDinterwordspacing}{\spaceskip=0pt\relax}
\providecommand{\BIBentryALTinterwordstretchfactor}{4}
\providecommand{\BIBentryALTinterwordspacing}{\spaceskip=\fontdimen2\font plus
\BIBentryALTinterwordstretchfactor\fontdimen3\font minus \fontdimen4\font\relax}
\providecommand{\BIBforeignlanguage}[2]{{%
\expandafter\ifx\csname l@#1\endcsname\relax
\typeout{** WARNING: IEEEtran.bst: No hyphenation pattern has been}%
\typeout{** loaded for the language `#1'. Using the pattern for}%
\typeout{** the default language instead.}%
\else
\language=\csname l@#1\endcsname
\fi
#2}}
\providecommand{\BIBdecl}{\relax}
\BIBdecl

\bibitem{DBLP:journals/tgrs/JiangZWWZYZCC22}
\BIBentryALTinterwordspacing
J.~Jiang, Q.~Zhang, W.~Wang, Y.~Wu, H.~Zheng, X.~Yao, Y.~Zhu, W.~Cao, and T.~Cheng, ``{MACA:} {A} relative radiometric correction method for multiflight unmanned aerial vehicle images based on concurrent satellite imagery,'' \emph{{IEEE} Trans. Geosci. Remote. Sens.}, vol.~60, pp. 1--14, 2022. [Online]. Available: \url{https://doi.org/10.1109/TGRS.2022.3158644}
\BIBentrySTDinterwordspacing

\bibitem{DBLP:journals/tgrs/WangGTGWGZH00CY24}
\BIBentryALTinterwordspacing
Y.~Wang, Y.~Gu, J.~Tang, B.~Guo, T.~A. Warner, C.~Guo, H.~Zheng, F.~Hosoi, T.~Cheng, Y.~Zhu, W.~Cao, and X.~Yao, ``Quantify wheat canopy leaf angle distribution using terrestrial laser scanning data,'' \emph{{IEEE} Trans. Geosci. Remote. Sens.}, vol.~62, pp. 1--15, 2024. [Online]. Available: \url{https://doi.org/10.1109/TGRS.2024.3353225}
\BIBentrySTDinterwordspacing

\bibitem{li2024investigation}
Q.~Li, Z.~Chai, R.~Yao, T.~Bai, and H.~Zhao, ``Investigation into uav applications for environmental ice detection and de-icing technology,'' \emph{Drones}, vol.~9, no.~1, p.~5, 2024.

\bibitem{xu2023survey}
H.~Xu, L.~Wang, W.~Han, Y.~Yang, J.~Li, Y.~Lu, and J.~Li, ``A survey on uav applications in smart city management: Challenges, advances, and opportunities,'' \emph{IEEE Journal of Selected Topics in Applied Earth Observations and Remote Sensing}, vol.~16, pp. 8982--9010, 2023.

\bibitem{guerin2021certifying}
J.~Gu{\'e}rin, K.~Delmas, and J.~Guiochet, ``Certifying emergency landing for safe urban uav,'' in \emph{2021 51st Annual IEEE/IFIP international conference on dependable systems and networks workshops (DSN-W)}.\hskip 1em plus 0.5em minus 0.4em\relax IEEE, 2021, pp. 55--62.

\bibitem{DBLP:conf/cvpr/Luo00LY024}
\BIBentryALTinterwordspacing
J.~Luo, X.~Yang, Y.~Yu, Q.~Li, J.~Yan, and Y.~Li, ``Pointobb: Learning oriented object detection via single point supervision,'' in \emph{{IEEE/CVF} Conference on Computer Vision and Pattern Recognition, {CVPR} 2024, Seattle, WA, USA, June 16-22, 2024}.\hskip 1em plus 0.5em minus 0.4em\relax {IEEE}, 2024, pp. 16\,730--16\,740. [Online]. Available: \url{https://doi.org/10.1109/CVPR52733.2024.01583}
\BIBentrySTDinterwordspacing

\bibitem{DBLP:conf/cvpr/Yu0LDD0Y24}
\BIBentryALTinterwordspacing
Y.~Yu, X.~Yang, Q.~Li, F.~Da, J.~Dai, Y.~Qiao, and J.~Yan, ``Point2rbox: Combine knowledge from synthetic visual patterns for end-to-end oriented object detection with single point supervision,'' in \emph{{IEEE/CVF} Conference on Computer Vision and Pattern Recognition, {CVPR} 2024, Seattle, WA, USA, June 16-22, 2024}.\hskip 1em plus 0.5em minus 0.4em\relax {IEEE}, 2024, pp. 16\,783--16\,793. [Online]. Available: \url{https://doi.org/10.1109/CVPR52733.2024.01588}
\BIBentrySTDinterwordspacing

\bibitem{DBLP:conf/cvpr/0010D23}
\BIBentryALTinterwordspacing
Y.~Yu and F.~Da, ``Phase-shifting coder: Predicting accurate orientation in oriented object detection,'' in \emph{{IEEE/CVF} Conference on Computer Vision and Pattern Recognition, {CVPR} 2023, Vancouver, BC, Canada, June 17-24, 2023}.\hskip 1em plus 0.5em minus 0.4em\relax {IEEE}, 2023, pp. 13\,354--13\,363. [Online]. Available: \url{https://doi.org/10.1109/CVPR52729.2023.01283}
\BIBentrySTDinterwordspacing

\bibitem{DBLP:conf/cvpr/HuaLLLZYB23}
\BIBentryALTinterwordspacing
W.~Hua, D.~Liang, J.~Li, X.~Liu, Z.~Zou, X.~Ye, and X.~Bai, ``{SOOD:} towards semi-supervised oriented object detection,'' in \emph{{IEEE/CVF} Conference on Computer Vision and Pattern Recognition, {CVPR} 2023, Vancouver, BC, Canada, June 17-24, 2023}.\hskip 1em plus 0.5em minus 0.4em\relax {IEEE}, 2023, pp. 15\,558--15\,567. [Online]. Available: \url{https://doi.org/10.1109/CVPR52729.2023.01493}
\BIBentrySTDinterwordspacing

\bibitem{DBLP:journals/pami/ChengYYYZXH23}
\BIBentryALTinterwordspacing
G.~Cheng, X.~Yuan, X.~Yao, K.~Yan, Q.~Zeng, X.~Xie, and J.~Han, ``Towards large-scale small object detection: Survey and benchmarks,'' \emph{{IEEE} Trans. Pattern Anal. Mach. Intell.}, vol.~45, no.~11, pp. 13\,467--13\,488, 2023. [Online]. Available: \url{https://doi.org/10.1109/TPAMI.2023.3290594}
\BIBentrySTDinterwordspacing

\bibitem{DBLP:conf/cvpr/XiaBDZBLDPZ18}
\BIBentryALTinterwordspacing
G.~Xia, X.~Bai, J.~Ding, Z.~Zhu, S.~J. Belongie, J.~Luo, M.~Datcu, M.~Pelillo, and L.~Zhang, ``{DOTA:} {A} large-scale dataset for object detection in aerial images,'' in \emph{2018 {IEEE} Conference on Computer Vision and Pattern Recognition, {CVPR} 2018, Salt Lake City, UT, USA, June 18-22, 2018}.\hskip 1em plus 0.5em minus 0.4em\relax Computer Vision Foundation / {IEEE} Computer Society, 2018, pp. 3974--3983. [Online]. Available: \url{http://openaccess.thecvf.com/content\_cvpr\_2018/html/Xia\_DOTA\_A\_Large-Scale\_CVPR\_2018\_paper.html}
\BIBentrySTDinterwordspacing

\bibitem{DBLP:journals/tgrs/ChengWLXLYH22}
\BIBentryALTinterwordspacing
G.~Cheng, J.~Wang, K.~Li, X.~Xie, C.~Lang, Y.~Yao, and J.~Han, ``Anchor-free oriented proposal generator for object detection,'' \emph{{IEEE} Trans. Geosci. Remote. Sens.}, vol.~60, pp. 1--11, 2022. [Online]. Available: \url{https://doi.org/10.1109/TGRS.2022.3183022}
\BIBentrySTDinterwordspacing

\bibitem{DBLP:conf/icpram/LiuYWY17}
\BIBentryALTinterwordspacing
Z.~Liu, L.~Yuan, L.~Weng, and Y.~Yang, ``A high resolution optical satellite image dataset for ship recognition and some new baselines,'' in \emph{Proceedings of the 6th International Conference on Pattern Recognition Applications and Methods, {ICPRAM} 2017, Porto, Portugal, February 24-26, 2017}, M.~D. Marsico, G.~S. di~Baja, and A.~L.~N. Fred, Eds.\hskip 1em plus 0.5em minus 0.4em\relax SciTePress, 2017, pp. 324--331. [Online]. Available: \url{https://doi.org/10.5220/0006120603240331}
\BIBentrySTDinterwordspacing

\bibitem{DBLP:journals/corr/abs-2003-06832}
\BIBentryALTinterwordspacing
K.~Chen, M.~Wu, J.~Liu, and C.~Zhang, ``{FGSD:} {A} dataset for fine-grained ship detection in high resolution satellite images,'' \emph{CoRR}, vol. abs/2003.06832, 2020. [Online]. Available: \url{https://arxiv.org/abs/2003.06832}
\BIBentrySTDinterwordspacing

\bibitem{DBLP:journals/jvcir/RazakarivonyJ16}
\BIBentryALTinterwordspacing
S.~Razakarivony and F.~Jurie, ``Vehicle detection in aerial imagery : {A} small target detection benchmark,'' \emph{J. Vis. Commun. Image Represent.}, vol.~34, pp. 187--203, 2016. [Online]. Available: \url{https://doi.org/10.1016/j.jvcir.2015.11.002}
\BIBentrySTDinterwordspacing

\bibitem{DBLP:conf/icip/ZhuCDFYJ15}
\BIBentryALTinterwordspacing
H.~Zhu, X.~Chen, W.~Dai, K.~Fu, Q.~Ye, and J.~Jiao, ``Orientation robust object detection in aerial images using deep convolutional neural network,'' in \emph{2015 {IEEE} International Conference on Image Processing, {ICIP} 2015, Quebec City, QC, Canada, September 27-30, 2015}.\hskip 1em plus 0.5em minus 0.4em\relax {IEEE}, 2015, pp. 3735--3739. [Online]. Available: \url{https://doi.org/10.1109/ICIP.2015.7351502}
\BIBentrySTDinterwordspacing

\bibitem{DBLP:conf/iccvw/DuZWWSLBSZKLZAS19}
\BIBentryALTinterwordspacing
D.~Du, Y.~Zhang, Z.~Wang, Z.~Wang, Z.~Song, Z.~Liu, L.~Bo, H.~Shi, R.~Zhu, A.~Kumar, A.~Li, A.~Zinollayev, A.~Askergaliyev, A.~Schumann, B.~Mao, P.~Zhu, B.~Lee, C.~Liu, C.~Chen, C.~Pan, C.~Huo, D.~Yu, D.~Cong, D.~Zeng, D.~R. Pailla, D.~Li, L.~Wen, D.~Wang, D.~Cho, D.~Zhang, F.~Bai, G.~Jose, G.~Gao, G.~Liu, H.~Xiong, H.~Qi, H.~Wang, X.~Bian, H.~Qiu, H.~Li, H.~Lu, I.~Kim, J.~Kim, J.~Shen, J.~Lee, J.~Ge, J.~Xu, J.~Zhou, H.~Ling, J.~Meier, J.~W. Choi, J.~Hu, J.~Zhang, J.~Huang, K.~Huang, K.~Wang, L.~Sommer, L.~Jin, L.~Zhang, Q.~Hu, L.~Huang, L.~Sun, L.~Steinmann, M.~Jia, N.~Xu, P.~Zhang, Q.~Chen, Q.~Lv, Q.~Liu, Q.~Cheng, T.~Peng, S.~S. Chennamsetty, S.~Chen, S.~Wei, S.~S.~S. Kruthiventi, S.~Hong, S.~Kang, T.~Wu, T.~Feng, V.~A. Kollerathu, W.~Li, J.~Zheng, W.~Dai, W.~Qin, W.~Wang, X.~Wang, X.~Chen, X.~Chen, X.~Sun, X.~Zhang, X.~Zhao, X.~Zhang, X.~Wang, X.~Zhang, X.~Chen, X.~Wei, X.~Zhang, Y.~Li, Y.~Chen, Y.~H. Toh, Y.~Zhang, Y.~Zhu, and Y.~Zhong, ``Visdrone-det2019: The vision meets drone object detection in image
  challenge results,'' in \emph{2019 {IEEE/CVF} International Conference on Computer Vision Workshops, {ICCV} Workshops 2019, Seoul, Korea (South), October 27-28, 2019}.\hskip 1em plus 0.5em minus 0.4em\relax {IEEE}, 2019, pp. 213--226. [Online]. Available: \url{https://doi.org/10.1109/ICCVW.2019.00030}
\BIBentrySTDinterwordspacing

\bibitem{DBLP:conf/icra/BozcanK20}
\BIBentryALTinterwordspacing
I.~Bozcan and E.~Kayacan, ``{AU-AIR:} {A} multi-modal unmanned aerial vehicle dataset for low altitude traffic surveillance,'' in \emph{2020 {IEEE} International Conference on Robotics and Automation, {ICRA} 2020, Paris, France, May 31 - August 31, 2020}.\hskip 1em plus 0.5em minus 0.4em\relax {IEEE}, 2020, pp. 8504--8510. [Online]. Available: \url{https://doi.org/10.1109/ICRA40945.2020.9196845}
\BIBentrySTDinterwordspacing

\bibitem{DBLP:conf/eccv/DuQYYDLZHT18}
\BIBentryALTinterwordspacing
D.~Du, Y.~Qi, H.~Yu, Y.~Yang, K.~Duan, G.~Li, W.~Zhang, Q.~Huang, and Q.~Tian, ``The unmanned aerial vehicle benchmark: Object detection and tracking,'' in \emph{Computer Vision - {ECCV} 2018 - 15th European Conference, Munich, Germany, September 8-14, 2018, Proceedings, Part {X}}, ser. Lecture Notes in Computer Science, V.~Ferrari, M.~Hebert, C.~Sminchisescu, and Y.~Weiss, Eds., vol. 11214.\hskip 1em plus 0.5em minus 0.4em\relax Springer, 2018, pp. 375--391. [Online]. Available: \url{https://doi.org/10.1007/978-3-030-01249-6\_23}
\BIBentrySTDinterwordspacing

\bibitem{DBLP:conf/iccv/HsiehLH17}
\BIBentryALTinterwordspacing
M.~Hsieh, Y.~Lin, and W.~H. Hsu, ``Drone-based object counting by spatially regularized regional proposal network,'' in \emph{{IEEE} International Conference on Computer Vision, {ICCV} 2017, Venice, Italy, October 22-29, 2017}.\hskip 1em plus 0.5em minus 0.4em\relax {IEEE} Computer Society, 2017, pp. 4165--4173. [Online]. Available: \url{https://doi.ieeecomputersociety.org/10.1109/ICCV.2017.446}
\BIBentrySTDinterwordspacing

\bibitem{DBLP:journals/tcsv/SunCZH22}
\BIBentryALTinterwordspacing
Y.~Sun, B.~Cao, P.~Zhu, and Q.~Hu, ``Drone-based rgb-infrared cross-modality vehicle detection via uncertainty-aware learning,'' \emph{{IEEE} Trans. Circuits Syst. Video Technol.}, vol.~32, no.~10, pp. 6700--6713, 2022. [Online]. Available: \url{https://doi.org/10.1109/TCSVT.2022.3168279}
\BIBentrySTDinterwordspacing

\bibitem{DBLP:journals/ijon/ZhouFLHP22}
\BIBentryALTinterwordspacing
J.~Zhou, K.~Feng, W.~Li, J.~Han, and F.~Pan, ``Ts\({}^{\mbox{4}}\)net: Two-stage sample selective strategy for rotating object detection,'' \emph{Neurocomputing}, vol. 501, pp. 753--764, 2022. [Online]. Available: \url{https://doi.org/10.1016/j.neucom.2022.06.049}
\BIBentrySTDinterwordspacing

\bibitem{DBLP:journals/staeors/HaroonSF20}
\BIBentryALTinterwordspacing
M.~Haroon, M.~Shahzad, and M.~M. Fraz, ``Multisized object detection using spaceborne optical imagery,'' \emph{{IEEE} J. Sel. Top. Appl. Earth Obs. Remote. Sens.}, vol.~13, pp. 3032--3046, 2020. [Online]. Available: \url{https://doi.org/10.1109/JSTARS.2020.3000317}
\BIBentrySTDinterwordspacing

\bibitem{DBLP:conf/eccv/MundhenkKSB16}
\BIBentryALTinterwordspacing
T.~N. Mundhenk, G.~Konjevod, W.~A. Sakla, and K.~Boakye, ``A large contextual dataset for classification, detection and counting of cars with deep learning,'' in \emph{Computer Vision - {ECCV} 2016 - 14th European Conference, Amsterdam, The Netherlands, October 11-14, 2016, Proceedings, Part {III}}, ser. Lecture Notes in Computer Science, B.~Leibe, J.~Matas, N.~Sebe, and M.~Welling, Eds., vol. 9907.\hskip 1em plus 0.5em minus 0.4em\relax Springer, 2016, pp. 785--800. [Online]. Available: \url{https://doi.org/10.1007/978-3-319-46487-9\_48}
\BIBentrySTDinterwordspacing

\bibitem{DBLP:journals/tip/ChengHZX19}
\BIBentryALTinterwordspacing
G.~Cheng, J.~Han, P.~Zhou, and D.~Xu, ``Learning rotation-invariant and fisher discriminative convolutional neural networks for object detection,'' \emph{{IEEE} Trans. Image Process.}, vol.~28, no.~1, pp. 265--278, 2019. [Online]. Available: \url{https://doi.org/10.1109/TIP.2018.2867198}
\BIBentrySTDinterwordspacing

\bibitem{xiao2015elliptic}
Z.~Xiao, Q.~Liu, G.~Tang, and X.~Zhai, ``Elliptic fourier transformation-based histograms of oriented gradients for rotationally invariant object detection in remote-sensing images,'' \emph{International Journal of Remote Sensing}, vol.~36, no.~2, pp. 618--644, 2015.

\bibitem{DBLP:journals/corr/abs-1909-00133}
\BIBentryALTinterwordspacing
K.~Li, G.~Wan, G.~Cheng, L.~Meng, and J.~Han, ``Object detection in optical remote sensing images: {A} survey and {A} new benchmark,'' \emph{CoRR}, vol. abs/1909.00133, 2019. [Online]. Available: \url{http://arxiv.org/abs/1909.00133}
\BIBentrySTDinterwordspacing

\bibitem{DBLP:journals/corr/abs-1802-07856}
\BIBentryALTinterwordspacing
D.~Lam, R.~Kuzma, K.~McGee, S.~Dooley, M.~Laielli, M.~Klaric, Y.~Bulatov, and B.~McCord, ``xview: Objects in context in overhead imagery,'' \emph{CoRR}, vol. abs/1802.07856, 2018. [Online]. Available: \url{http://arxiv.org/abs/1802.07856}
\BIBentrySTDinterwordspacing

\bibitem{DBLP:journals/tgrs/ZhangYFL19}
\BIBentryALTinterwordspacing
Y.~Zhang, Y.~Yuan, Y.~Feng, and X.~Lu, ``Hierarchical and robust convolutional neural network for very high-resolution remote sensing object detection,'' \emph{{IEEE} Trans. Geosci. Remote. Sens.}, vol.~57, no.~8, pp. 5535--5548, 2019. [Online]. Available: \url{https://doi.org/10.1109/TGRS.2019.2900302}
\BIBentrySTDinterwordspacing

\bibitem{DBLP:journals/tip/ZouS18}
\BIBentryALTinterwordspacing
Z.~Zou and Z.~Shi, ``Random access memories: {A} new paradigm for target detection in high resolution aerial remote sensing images,'' \emph{{IEEE} Trans. Image Process.}, vol.~27, no.~3, pp. 1100--1111, 2018. [Online]. Available: \url{https://doi.org/10.1109/TIP.2017.2773199}
\BIBentrySTDinterwordspacing

\bibitem{DBLP:conf/icpr/WangYGZX20}
\BIBentryALTinterwordspacing
J.~Wang, W.~Yang, H.~Guo, R.~Zhang, and G.~Xia, ``Tiny object detection in aerial images,'' in \emph{25th International Conference on Pattern Recognition, {ICPR} 2020, Virtual Event / Milan, Italy, January 10-15, 2021}.\hskip 1em plus 0.5em minus 0.4em\relax {IEEE}, 2020, pp. 3791--3798. [Online]. Available: \url{https://doi.org/10.1109/ICPR48806.2021.9413340}
\BIBentrySTDinterwordspacing

\bibitem{DBLP:journals/corr/abs-2103-05569}
\BIBentryALTinterwordspacing
X.~Sun, P.~Wang, Z.~Yan, F.~Xu, R.~Wang, W.~Diao, J.~Chen, J.~Li, Y.~Feng, T.~Xu, M.~Weinmann, S.~Hinz, C.~Wang, and K.~Fu, ``{FAIR1M:} {A} benchmark dataset for fine-grained object recognition in high-resolution remote sensing imagery,'' \emph{CoRR}, vol. abs/2103.05569, 2021. [Online]. Available: \url{https://arxiv.org/abs/2103.05569}
\BIBentrySTDinterwordspacing

\bibitem{DBLP:conf/icpr/AzimiBHK20}
\BIBentryALTinterwordspacing
S.~M. Azimi, R.~Bahmanyar, C.~Henry, and F.~Kurz, ``{EAGLE:} large-scale vehicle detection dataset in real-world scenarios using aerial imagery,'' in \emph{25th International Conference on Pattern Recognition, {ICPR} 2020, Virtual Event / Milan, Italy, January 10-15, 2021}.\hskip 1em plus 0.5em minus 0.4em\relax {IEEE}, 2020, pp. 6920--6927. [Online]. Available: \url{https://doi.org/10.1109/ICPR48806.2021.9412353}
\BIBentrySTDinterwordspacing

\bibitem{DBLP:journals/corr/abs-2409-19833}
\BIBentryALTinterwordspacing
C.~Feng, Z.~Chen, R.~Kou, G.~Gao, C.~Wang, X.~Li, X.~Shu, Y.~Dai, Q.~Fu, and J.~Yang, ``Hazydet: Open-source benchmark for drone-view object detection with depth-cues in hazy scenes,'' \emph{CoRR}, vol. abs/2409.19833, 2024. [Online]. Available: \url{https://doi.org/10.48550/arXiv.2409.19833}
\BIBentrySTDinterwordspacing

\bibitem{DBLP:conf/nips/YangYYMWTY21}
\BIBentryALTinterwordspacing
X.~Yang, X.~Yang, J.~Yang, Q.~Ming, W.~Wang, Q.~Tian, and J.~Yan, ``Learning high-precision bounding box for rotated object detection via kullback-leibler divergence,'' in \emph{Advances in Neural Information Processing Systems 34: Annual Conference on Neural Information Processing Systems 2021, NeurIPS 2021, December 6-14, 2021, virtual}, M.~Ranzato, A.~Beygelzimer, Y.~N. Dauphin, P.~Liang, and J.~W. Vaughan, Eds., 2021, pp. 18\,381--18\,394. [Online]. Available: \url{https://proceedings.neurips.cc/paper/2021/hash/98f13708210194c475687be6106a3b84-Abstract.html}
\BIBentrySTDinterwordspacing

\bibitem{DBLP:journals/corr/abs-2303-04989}
\BIBentryALTinterwordspacing
Y.~Zeng, X.~Yang, Q.~Li, Y.~Chen, and J.~Yan, ``{ARS-DETR:} aspect ratio sensitive oriented object detection with transformer,'' \emph{CoRR}, vol. abs/2303.04989, 2023. [Online]. Available: \url{https://doi.org/10.48550/arXiv.2303.04989}
\BIBentrySTDinterwordspacing

\bibitem{DBLP:journals/pami/RenHG017}
\BIBentryALTinterwordspacing
S.~Ren, K.~He, R.~B. Girshick, and J.~Sun, ``Faster {R-CNN:} towards real-time object detection with region proposal networks,'' \emph{{IEEE} Trans. Pattern Anal. Mach. Intell.}, vol.~39, no.~6, pp. 1137--1149, 2017. [Online]. Available: \url{https://doi.org/10.1109/TPAMI.2016.2577031}
\BIBentrySTDinterwordspacing

\bibitem{DBLP:conf/iccv/LinGGHD17}
\BIBentryALTinterwordspacing
T.~Lin, P.~Goyal, R.~B. Girshick, K.~He, and P.~Doll{\'{a}}r, ``Focal loss for dense object detection,'' in \emph{{IEEE} International Conference on Computer Vision, {ICCV} 2017, Venice, Italy, October 22-29, 2017}.\hskip 1em plus 0.5em minus 0.4em\relax {IEEE} Computer Society, 2017, pp. 2999--3007. [Online]. Available: \url{https://doi.org/10.1109/ICCV.2017.324}
\BIBentrySTDinterwordspacing

\bibitem{DBLP:conf/iccv/YangLHWL19}
\BIBentryALTinterwordspacing
Z.~Yang, S.~Liu, H.~Hu, L.~Wang, and S.~Lin, ``Reppoints: Point set representation for object detection,'' in \emph{2019 {IEEE/CVF} International Conference on Computer Vision, {ICCV} 2019, Seoul, Korea (South), October 27 - November 2, 2019}.\hskip 1em plus 0.5em minus 0.4em\relax {IEEE}, 2019, pp. 9656--9665. [Online]. Available: \url{https://doi.org/10.1109/ICCV.2019.00975}
\BIBentrySTDinterwordspacing

\bibitem{DBLP:conf/cvpr/DingXLXL19}
\BIBentryALTinterwordspacing
J.~Ding, N.~Xue, Y.~Long, G.~Xia, and Q.~Lu, ``Learning roi transformer for oriented object detection in aerial images,'' in \emph{{IEEE} Conference on Computer Vision and Pattern Recognition, {CVPR} 2019, Long Beach, CA, USA, June 16-20, 2019}.\hskip 1em plus 0.5em minus 0.4em\relax Computer Vision Foundation / {IEEE}, 2019, pp. 2849--2858. [Online]. Available: \url{http://openaccess.thecvf.com/content\_CVPR\_2019/html/Ding\_Learning\_RoI\_Transformer\_for\_Oriented\_Object\_Detection\_in\_Aerial\_Images\_CVPR\_2019\_paper.html}
\BIBentrySTDinterwordspacing

\bibitem{DBLP:conf/eccv/0005Y20}
\BIBentryALTinterwordspacing
X.~Yang and J.~Yan, ``Arbitrary-oriented object detection with circular smooth label,'' in \emph{Computer Vision - {ECCV} 2020 - 16th European Conference, Glasgow, UK, August 23-28, 2020, Proceedings, Part {VIII}}, ser. Lecture Notes in Computer Science, A.~Vedaldi, H.~Bischof, T.~Brox, and J.~Frahm, Eds., vol. 12353.\hskip 1em plus 0.5em minus 0.4em\relax Springer, 2020, pp. 677--694. [Online]. Available: \url{https://doi.org/10.1007/978-3-030-58598-3\_40}
\BIBentrySTDinterwordspacing

\bibitem{DBLP:conf/cvpr/ZhangCYLL20}
\BIBentryALTinterwordspacing
S.~Zhang, C.~Chi, Y.~Yao, Z.~Lei, and S.~Z. Li, ``Bridging the gap between anchor-based and anchor-free detection via adaptive training sample selection,'' in \emph{2020 {IEEE/CVF} Conference on Computer Vision and Pattern Recognition, {CVPR} 2020, Seattle, WA, USA, June 13-19, 2020}.\hskip 1em plus 0.5em minus 0.4em\relax Computer Vision Foundation / {IEEE}, 2020, pp. 9756--9765. [Online]. Available: \url{https://openaccess.thecvf.com/content\_CVPR\_2020/html/Zhang\_Bridging\_the\_Gap\_Between\_Anchor-Based\_and\_Anchor-Free\_Detection\_via\_Adaptive\_CVPR\_2020\_paper.html}
\BIBentrySTDinterwordspacing

\bibitem{DBLP:journals/pami/XuFWWCXB21}
\BIBentryALTinterwordspacing
Y.~Xu, M.~Fu, Q.~Wang, Y.~Wang, K.~Chen, G.~Xia, and X.~Bai, ``Gliding vertex on the horizontal bounding box for multi-oriented object detection,'' \emph{{IEEE} Trans. Pattern Anal. Mach. Intell.}, vol.~43, no.~4, pp. 1452--1459, 2021. [Online]. Available: \url{https://doi.org/10.1109/TPAMI.2020.2974745}
\BIBentrySTDinterwordspacing

\bibitem{DBLP:conf/iccv/Xie0WYH21}
\BIBentryALTinterwordspacing
X.~Xie, G.~Cheng, J.~Wang, X.~Yao, and J.~Han, ``Oriented {R-CNN} for object detection,'' in \emph{2021 {IEEE/CVF} International Conference on Computer Vision, {ICCV} 2021, Montreal, QC, Canada, October 10-17, 2021}.\hskip 1em plus 0.5em minus 0.4em\relax {IEEE}, 2021, pp. 3500--3509. [Online]. Available: \url{https://doi.org/10.1109/ICCV48922.2021.00350}
\BIBentrySTDinterwordspacing

\bibitem{DBLP:conf/icml/0005YMWZ021}
\BIBentryALTinterwordspacing
X.~Yang, J.~Yan, Q.~Ming, W.~Wang, X.~Zhang, and Q.~Tian, ``Rethinking rotated object detection with gaussian wasserstein distance loss,'' in \emph{Proceedings of the 38th International Conference on Machine Learning, {ICML} 2021, 18-24 July 2021, Virtual Event}, ser. Proceedings of Machine Learning Research, M.~Meila and T.~Zhang, Eds., vol. 139.\hskip 1em plus 0.5em minus 0.4em\relax {PMLR}, 2021, pp. 11\,830--11\,841. [Online]. Available: \url{http://proceedings.mlr.press/v139/yang21l.html}
\BIBentrySTDinterwordspacing

\bibitem{DBLP:journals/tgrs/HanDLX22}
\BIBentryALTinterwordspacing
J.~Han, J.~Ding, J.~Li, and G.~Xia, ``Align deep features for oriented object detection,'' \emph{{IEEE} Trans. Geosci. Remote. Sens.}, vol.~60, pp. 1--11, 2022. [Online]. Available: \url{https://doi.org/10.1109/TGRS.2021.3062048}
\BIBentrySTDinterwordspacing

\bibitem{DBLP:conf/aaai/YangYFH21}
\BIBentryALTinterwordspacing
X.~Yang, J.~Yan, Z.~Feng, and T.~He, ``R3det: Refined single-stage detector with feature refinement for rotating object,'' in \emph{Thirty-Fifth {AAAI} Conference on Artificial Intelligence, {AAAI} 2021, Thirty-Third Conference on Innovative Applications of Artificial Intelligence, {IAAI} 2021, The Eleventh Symposium on Educational Advances in Artificial Intelligence, {EAAI} 2021, Virtual Event, February 2-9, 2021}.\hskip 1em plus 0.5em minus 0.4em\relax {AAAI} Press, 2021, pp. 3163--3171. [Online]. Available: \url{https://doi.org/10.1609/aaai.v35i4.16426}
\BIBentrySTDinterwordspacing

\bibitem{DBLP:conf/cvpr/HanD0X21}
\BIBentryALTinterwordspacing
J.~Han, J.~Ding, N.~Xue, and G.~Xia, ``Redet: {A} rotation-equivariant detector for aerial object detection,'' in \emph{{IEEE} Conference on Computer Vision and Pattern Recognition, {CVPR} 2021, virtual, June 19-25, 2021}.\hskip 1em plus 0.5em minus 0.4em\relax Computer Vision Foundation / {IEEE}, 2021, pp. 2786--2795. [Online]. Available: \url{https://openaccess.thecvf.com/content/CVPR2021/html/Han\_ReDet\_A\_Rotation-Equivariant\_Detector\_for\_Aerial\_Object\_Detection\_CVPR\_2021\_paper.html}
\BIBentrySTDinterwordspacing

\bibitem{DBLP:conf/cvpr/LiCHZ22}
\BIBentryALTinterwordspacing
W.~Li, Y.~Chen, K.~Hu, and J.~Zhu, ``Oriented reppoints for aerial object detection,'' in \emph{{IEEE/CVF} Conference on Computer Vision and Pattern Recognition, {CVPR} 2022, New Orleans, LA, USA, June 18-24, 2022}.\hskip 1em plus 0.5em minus 0.4em\relax {IEEE}, 2022, pp. 1819--1828. [Online]. Available: \url{https://doi.org/10.1109/CVPR52688.2022.00187}
\BIBentrySTDinterwordspacing

\bibitem{DBLP:conf/iclr/00050ZYWY0023}
\BIBentryALTinterwordspacing
X.~Yang, Y.~Zhou, G.~Zhang, J.~Yang, W.~Wang, J.~Yan, X.~Zhang, and Q.~Tian, ``The kfiou loss for rotated object detection,'' in \emph{The Eleventh International Conference on Learning Representations, {ICLR} 2023, Kigali, Rwanda, May 1-5, 2023}.\hskip 1em plus 0.5em minus 0.4em\relax OpenReview.net, 2023. [Online]. Available: \url{https://openreview.net/forum?id=qUKsCztWlKq}
\BIBentrySTDinterwordspacing

\bibitem{DBLP:conf/iclr/0005ZL0WY23}
\BIBentryALTinterwordspacing
X.~Yang, G.~Zhang, W.~Li, Y.~Zhou, X.~Wang, and J.~Yan, ``H2rbox: Horizontal box annotation is all you need for oriented object detection,'' in \emph{The Eleventh International Conference on Learning Representations, {ICLR} 2023, Kigali, Rwanda, May 1-5, 2023}.\hskip 1em plus 0.5em minus 0.4em\relax OpenReview.net, 2023. [Online]. Available: \url{https://openreview.net/forum?id=NPfDKT9OUJ3}
\BIBentrySTDinterwordspacing

\bibitem{DBLP:conf/cvpr/XuDWYYYX23}
\BIBentryALTinterwordspacing
C.~Xu, J.~Ding, J.~Wang, W.~Yang, H.~Yu, L.~Yu, and G.~Xia, ``Dynamic coarse-to-fine learning for oriented tiny object detection,'' in \emph{{IEEE/CVF} Conference on Computer Vision and Pattern Recognition, {CVPR} 2023, Vancouver, BC, Canada, June 17-24, 2023}.\hskip 1em plus 0.5em minus 0.4em\relax {IEEE}, 2023, pp. 7318--7328. [Online]. Available: \url{https://doi.org/10.1109/CVPR52729.2023.00707}
\BIBentrySTDinterwordspacing

\bibitem{DBLP:conf/iccv/LiHZCYL23}
\BIBentryALTinterwordspacing
Y.~Li, Q.~Hou, Z.~Zheng, M.~Cheng, J.~Yang, and X.~Li, ``Large selective kernel network for remote sensing object detection,'' in \emph{{IEEE/CVF} International Conference on Computer Vision, {ICCV} 2023, Paris, France, October 1-6, 2023}.\hskip 1em plus 0.5em minus 0.4em\relax {IEEE}, 2023, pp. 16\,748--16\,759. [Online]. Available: \url{https://doi.org/10.1109/ICCV51070.2023.01540}
\BIBentrySTDinterwordspacing

\bibitem{DBLP:conf/nips/YuYL0DY23}
\BIBentryALTinterwordspacing
Y.~Yu, X.~Yang, Q.~Li, Y.~Zhou, F.~Da, and J.~Yan, ``H2rbox-v2: Incorporating symmetry for boosting horizontal box supervised oriented object detection,'' in \emph{Advances in Neural Information Processing Systems 36: Annual Conference on Neural Information Processing Systems 2023, NeurIPS 2023, New Orleans, LA, USA, December 10 - 16, 2023}, A.~Oh, T.~Naumann, A.~Globerson, K.~Saenko, M.~Hardt, and S.~Levine, Eds., 2023. [Online]. Available: \url{http://papers.nips.cc/paper\_files/paper/2023/hash/b9603de9e49d0838e53b6c9cf9d06556-Abstract-Conference.html}
\BIBentrySTDinterwordspacing

\bibitem{DBLP:journals/tgrs/ZhaoDZZDYE24}
\BIBentryALTinterwordspacing
J.~Zhao, Z.~Ding, Y.~Zhou, H.~Zhu, W.~Du, R.~Yao, and A.~El{-}Saddik, ``Orientedformer: An end-to-end transformer-based oriented object detector in remote sensing images,'' \emph{{IEEE} Trans. Geosci. Remote. Sens.}, vol.~62, pp. 1--16, 2024. [Online]. Available: \url{https://doi.org/10.1109/TGRS.2024.3456240}
\BIBentrySTDinterwordspacing

\bibitem{DBLP:journals/tgrs/LuCSTL24}
\BIBentryALTinterwordspacing
W.~Lu, S.~Chen, Q.~Shu, J.~Tang, and B.~Luo, ``Decouplenet: {A} lightweight backbone network with efficient feature decoupling for remote sensing visual tasks,'' \emph{{IEEE} Trans. Geosci. Remote. Sens.}, vol.~62, pp. 1--13, 2024. [Online]. Available: \url{https://doi.org/10.1109/TGRS.2024.3465496}
\BIBentrySTDinterwordspacing

\end{thebibliography}






\begin{IEEEbiographynophoto}{Kai Ye}
received the B.S. degree in Artificial Intelligence from Xiamen University, China. He is currently pursuing the Ph.D. degree at the MAC Lab, School of Informatics, Xiamen University, China. His research interests include UAV Perception, open-vocabulary detection and domain incremental learning.
\end{IEEEbiographynophoto}

\begin{IEEEbiographynophoto}{Bowen Liu}
is currently a senior student majoring in Computer Science and Technology at the School of Information, Xiamen University. He joined the MAC Laboratory at the School of Information, Xiamen University in September 2024. He has been recommended for graduate studies and continues to pursue a master's degree at Xiamen University.
\end{IEEEbiographynophoto}

\begin{IEEEbiographynophoto}{Haidi Tang}
is currently a senior student majoring in Computer Science and Technology at the School of Information, Xiamen University. He joined the MAC Laboratory at the School of Information, Xiamen University in September 2024. He has been recommended for graduate studies and continues to pursue a master's degree at Xiamen University.
\end{IEEEbiographynophoto}

\begin{IEEEbiographynophoto}{Pingyang Dai}
 received the M.S. degree in computer science and the Ph.D. degree in automation from
 Xiamen University, Xiamen, China, in 2003 and 2013, respectively.
 
 He is currently a Senior Engineer with the Key Laboratory of Multimedia Trusted Perception and Efficient Computing, Ministry of Education of China, and the School of Informatics, Xiamen University. His research interests include computer vision and machine learning.
\end{IEEEbiographynophoto}

\begin{IEEEbiographynophoto}{Liujuan Cao}
(Member, IEEE) received the B.S., M.S., and Ph.D. degrees from the School of Com
puter Science and Technology, Harbin Engineering University, Harbin, China, in 2005, 2008, and 2013, respectively.

She is currently an Associate Professor with Xiamen University, Xiamen, China. She has authored over 40 papers in the top and major tired journals and conferences, including Conference on Computer Vision and Pattern Recognition (CVPR) and IEEE TRANSACTIONS ON IMAGE PROCESSING (TIP).Her research interests include computer vision and pattern recognition.

Dr. Cao is also the Financial Chair for the IEEE International Workshop on Multimedia Signal Processing (MMSP) 2015, the Workshop Chair for the ACM International Conference on Internet Multimedia Computing and Service (ICIMCS) 2016, and the Local Chair for the Visual and Learning Seminar 2017.
\end{IEEEbiographynophoto}

\begin{IEEEbiographynophoto}{Rongrong Ji}
(Senior Member, IEEE) is currently a Nanqiang Distinguished Professor at Xiamen University, Xiamen, China; the Deputy Director of the Office of Science and Technology, Xiamen University; and the Director of the Media Analytics and Computing Laboratory, Xiamen University. He has authored 50+ articles in ACM/IEEE TRANSACTIONS, including the IEEE TRANSACTIONS ON PATTERN ANALYSIS AND MACHINE INTELLIGENCE (TPAMI) and International Journal of Computer Vision (IJCV), and 100+ full papers on top-tier conferences, such as Conference on Computer Vision and Pattern Recognition (CVPR) and Advances in Neural Information Processing Systems (NeurIPS). His research interests include computer vision, multimedia analysis, and machine learning.

Dr. Ji is also an Advisory Member for Artificial Intelligence Construction in the Electronic Information Education Committee of the National Ministry of Education. He was a recipient of the Best Paper Award of ACM Multimedia 2011. He was recognized as the National Science Foundation for Excellent Young Scholar in 2014, the National Ten Thousand Plan for Young Top Talent in 2017, and the National Science Foundation for Distinguished Young Scholar in 2020. He has served as the Area Chair for top-tier conferences, such as CVPR and ACM Multimedia. His publications have got over 20K citations in Google Scholar.
\end{IEEEbiographynophoto}

\vfill

\end{document}